\documentclass{article}

\usepackage{microtype}
\usepackage{graphicx}
\graphicspath{ {images/} }
\usepackage{subfigure}
\usepackage{booktabs} % for professional tables

\usepackage{hyperref}

\usepackage{amsmath}
\DeclareMathOperator{\Var}{Var}
\newcommand{\E}{\mathbb{E}}
\usepackage{cuted}
\usepackage{widetext}
\usepackage{xcolor}
\usepackage{amsfonts}
\usepackage[skip=0pt]{caption}

\usepackage[accepted]{icml2018}

\icmltitlerunning{The Mirage of Action-Dependent Baselines in Reinforcement Learning}

\begin{document}

\twocolumn[
\icmltitle{The Mirage of Action-Dependent Baselines in Reinforcement Learning}



\icmlsetsymbol{equal}{*}

\begin{icmlauthorlist}
\icmlauthor{George Tucker}{goo}
\icmlauthor{Surya Bhupatiraju}{goo,res}
\icmlauthor{Shixiang Gu}{goo,cam,mpi}
\icmlauthor{Richard E. Turner}{cam}
\icmlauthor{Zoubin Ghahramani}{cam,uber}
\icmlauthor{Sergey Levine}{goo,berk}
\end{icmlauthorlist}

\icmlaffiliation{goo}{Google Brain, USA}
\icmlaffiliation{cam}{University of Cambridge, UK}
\icmlaffiliation{mpi}{Max Planck Institute for Intelligent Systems, Germany}
\icmlaffiliation{berk}{UC Berkeley, USA}
\icmlaffiliation{uber}{Uber AI Labs, USA}
\icmlaffiliation{res}{Work was done during the Google AI Residency.}

\icmlcorrespondingauthor{George Tucker}{gjt@google.com}

\icmlkeywords{Machine Learning, Reinforcement Learning, Variance Reduction, Control Variate, Baseline, Action-dependent baseline}

\vskip 0.3in
]



\printAffiliationsAndNotice{}  % leave blank if no need to mention equal contribution

\begin{abstract}
%Model-free reinforcement learning with flexible function approximators has shown success in goal-directed sequential decision-making problems. 
Policy gradient methods are a widely used class of model-free reinforcement learning algorithms where a state-dependent baseline is used to reduce gradient estimator variance. Several recent papers extend the baseline to depend on both the state and action and suggest that this significantly reduces variance and improves sample efficiency without introducing bias into the gradient estimates. To better understand this development, we decompose the variance of the policy gradient estimator and numerically show that learned state-action-dependent baselines do not in fact reduce variance over a state-dependent baseline in commonly tested benchmark domains. We confirm this unexpected result by reviewing the open-source code accompanying these prior papers, and show that subtle implementation decisions cause deviations from the methods presented in the papers and explain the source of the previously observed empirical gains. Furthermore, the variance decomposition highlights areas for improvement, which we demonstrate by illustrating a simple change to the typical value function parameterization that can significantly improve performance.
\end{abstract}

\section{Introduction}
Model-free reinforcement learning (RL) with flexible function approximators, such as neural networks (i.e., deep reinforcement learning), has shown success in goal-directed sequential decision-making problems in high dimensional state spaces~\citep{mnih2015human,schulman2015high,lillicrap2015continuous,silver2016mastering}. Policy gradient methods~\citep{williams1992simple,sutton2000policy,kakade2002natural,peters2006policy,silver2014deterministic,schulman2015trust,schulman2017proximal} are a class of model-free RL algorithms that have found widespread adoption due to their stability and ease of use. Because these methods directly estimate the gradient of the expected reward RL objective, they exhibit stable convergence both in theory and practice~\citep{sutton2000policy,kakade2002natural,schulman2015trust,gu2017interpolated}.
In contrast, methods such as Q-learning lack convergence guarantees in the case of nonlinear function approximation~\citep{sutton1998reinforcement}.

On-policy Monte-Carlo policy gradient estimates suffer from high variance, and therefore require large batch sizes to reliably estimate the gradient for stable iterative optimization~\citep{schulman2015trust}. This limits their applicability to real-world problems, where sample efficiency is a critical constraint. Actor-critic methods~\citep{sutton2000policy,silver2014deterministic} and $\lambda$-weighted return estimation~\citep{tesauro1995temporal,schulman2015high} replace the high variance Monte-Carlo return with an estimate based on the sampled return and a function approximator. This reduces variance at the expense of introducing bias from the function approximator, which can lead to instability and sensitivity to hyperparameters. In contrast, state-dependent baselines~\citep{williams1992simple,weaver2001optimal} reduce variance without introducing bias. This is desirable because it does not compromise the stability of the original method.

\citet{gu2017q,grathwohl2018backpropagation,liu2018sample,wu2018variance} present promising results extending the classic state-dependent baselines to state-action-dependent baselines. The standard explanation for the benefits of such approaches is that they achieve large reductions in variance~\citep{grathwohl2018backpropagation,liu2018sample}, which translates to improvements over methods that only condition the baseline on the state. This line of investigation is attractive, because by definition, baselines do not introduce bias and thus do not compromise the stability of the underlying policy gradient algorithm, but still provide improved sample efficiency. In other words, they retain the advantages of the underlying algorithms with no unintended side-effects.

In this paper, we aim to improve our understanding of state-action-dependent baselines and to identify targets for further unbiased variance reduction. Toward this goal, we present a decomposition of the variance of the policy gradient estimator which isolates the potential variance reduction due to state-action-dependent baselines. We numerically evaluate the variance components on a synthetic linear-quadratic-Gaussian (LQG) task, where the variances are nearly analytically tractable, and on benchmark continuous control tasks and draw two conclusions: (1) on these tasks, a learned state-action-dependent baseline does not significantly reduce variance over a learned state-dependent baseline%, a conclusion seemingly at odds with previous work
, and (2) the variance caused by using a function approximator for the value function or state-dependent baseline is much larger than the variance reduction from adding action dependence to the baseline.

To resolve the apparent contradiction arising from (1), we carefully reviewed the open-source implementations\footnote{At the time of submission, code for \citep{wu2018variance} was not available.} accompanying Q-prop~\citep{gu2017q}, Stein control variates~\citep{liu2018sample}, and LAX~\citep{grathwohl2018backpropagation} and show that subtle implementation decisions cause the code to diverge from the unbiased methods presented in the papers. We explain and empirically evaluate variants of these prior methods to demonstrate that these subtle implementation details, which trade variance for bias, are in fact crucial for their empirical success. These results motivate further study of these design decisions.

The second observation (2), that function approximators poorly estimate the value function, suggests that there is room for improvement. Although many common benchmark tasks are finite horizon problems, most value function parameterizations ignore this fact. We propose a horizon-aware value function parameterization, and this improves performance compared with the state-action-dependent baseline without biasing the underlying method.

We emphasize that without the open-source code accompanying~\citep{gu2017q,liu2018sample,grathwohl2018backpropagation}, this work would not be possible. Releasing the code has allowed us to present a new view on their work and to identify interesting implementation decisions for further study that the original authors may not have been aware of. 

We have made our code and additional visualizations available at  \url{https://sites.google.com/view/mirage-rl}.

\section{Background}
Reinforcement learning aims to learn a policy for an agent to maximize a sum of reward signals~\citep{sutton1998reinforcement}. The agent starts at an initial state $s_0 \sim P(s_0)$. Then, the agent repeatedly samples an action $a_t$ from a policy $\pi_\theta(a_t | s_t)$ with parameters $\theta$, receives a reward $r_t \sim P(r_t | s_t, a_t)$, and transitions to a subsequent state $s_{t+1}$ according to the Markovian dynamics $P(s_{t+1} | a_t, s_t)$ of the environment. This generates a trajectory of states, actions, and rewards $(s_0, a_0, r_0, s_1, a_1, \ldots)$. We abbreviate the trajectory after the initial state and action by $\tau$.

The goal is to maximize the discounted sum of rewards along sampled trajectories
\[ J(\theta) = \mathbb{E}_{s_0, a_0, \tau} \left[ \sum_{t = 0}^\infty \gamma^t r_t \right] = \E_{s \sim \rho^\pi(s), a, \tau} \left[ \sum_{t=0}^\infty \gamma^t r_t \right], \]
where $\gamma \in [0, 1)$ is a discount parameter and $\rho^\pi(s) = \sum_{t = 0}^\infty \gamma^t P^\pi(s_t = s)$ is the unnormalized discounted state visitation frequency. 

Policy gradient methods differentiate the expected return objective with respect to the policy parameters and apply gradient-based optimization~\citep{sutton1998reinforcement}. The policy gradient can be written as an expectation amenable to Monte Carlo estimation
\begin{align}
\nabla_\theta J(\theta) &= \E_{s \sim \rho^\pi(s), a, \tau} \left[ Q^\pi(s, a) \nabla \log \pi(a | s) \right] \nonumber \\
	&= \E_{s \sim \rho^\pi(s), a, \tau} \left[ A^\pi(s, a) \nabla \log \pi(a | s) \right] \nonumber
\end{align}
where $Q^\pi(s, a) = \E_\tau \left[ \sum_{t = 0}^\infty \gamma^t r_t | s_0 = s, a_0 = a \right]$ is the state-action value function, $V^\pi(s) = \E_{a}\left[ Q^\pi(s, a) \right]$ is the value function, and $A^\pi(s, a) = Q^\pi(s, a) - V^\pi(s)$ is the advantage function. The equality in the last line follows from the fact that $\E_{a} \left[ \nabla \log \pi(a | s) \right] = 0$~\citep{williams1992simple}.

In practice, most policy gradient methods (including this paper) use the \emph{undiscounted} state visitation frequencies (i.e., $\gamma=1$ for $\rho^\pi(s)$), which produces a biased estimator for $\nabla J(\theta)$ and more closely aligns with maximizing average reward~\cite{thomas2014bias}.

We can estimate the gradient with a Monte-Carlo estimator
\begin{align}
	\hat{g}(s, a, \tau) = \hat{A}(s, a, \tau)\nabla \log \pi_\theta(a|s),
    \label{eq:policy-gradient}
\end{align}
where $\hat{A}$ is an estimator of the advantage function up to a state-dependent constant (e.g., $\sum_t \gamma^t r_t$).

\subsection{Advantage Function Estimation}
Given a value function estimator, $\hat{V}(s)$, we can form a $k$-step advantage function estimator,
\begin{align}
\hat{A}^{(k)}(s_t,a_t,\tau_{t+1}) = \sum_{i=0}^{k-1} \gamma^i r_{t+i} + \gamma^k \hat{V}(s_{t+k}) - \hat{V}(s_t) , \nonumber
\end{align}
where $k\in\{1,2,...,\infty\}$ and $\tau_{t+1} = (r_t, s_{t+1}, a_{t+1}, \ldots)$.
$\hat{A}^{(\infty)}(s_t,a_t,\tau_{t+1})$ produces an unbiased gradient estimator when used in Eq.~\ref{eq:policy-gradient} regardless of the choice of $\hat{V}(s)$. However, the other estimators ($k < \infty$) produce biased estimates unless $\hat{V}(s)=V^\pi(s)$. Advantage actor critic (A2C and A3C) methods~\cite{mnih2016asynchronous} and generalized advantage estimators (GAE)~\cite{schulman2015high} use a single or linear combination of $\hat{A}^{(k)}$ estimators as the advantage estimator in Eq.~\ref{eq:policy-gradient}. In practice, the value function estimator is never perfect, so these methods produce biased gradient estimates. As a result, the hyperparameters that control the combination of $\hat{A}^{(k)}$ must be carefully tuned to balance bias and variance~\citep{schulman2015high}, demonstrating the perils and sensitivity of biased gradient estimators. For the experiments in this paper, unless stated otherwise, we use the GAE estimator. Our focus will be on the additional bias introduced beyond that of GAE.

\subsection{Baselines for Variance Reduction}
The policy gradient estimator in Eq.~\ref{eq:policy-gradient} typically suffers from high variance. Control variates are a well-studied technique for reducing variance in Monte Carlo estimators without biasing the estimator~\citep{owen2013monte}. They require a correlated function whose expectation we can analytically evaluate or estimate with low variance. Because $\E_{a|s} \left[ \nabla \log \pi(a | s) \right] = 0$, any function of the form $\phi(s) \nabla \log \pi(a | s)$ can serve as a control variate, where $\phi(s)$ is commonly referred to as a baseline~\citep{williams1992simple}. With a baseline, the policy gradient estimator becomes
\begin{align*}
	\hat{g}(s, a, \tau) =& \left( \hat{A}(s, a, \tau) - \phi(s) \right) \nabla \log \pi(a|s),
\end{align*}
which does not introduce bias. Several recent methods~\citep{gu2017q,thomas2017policy,grathwohl2018backpropagation,liu2018sample,wu2018variance} have extended the approach to state-action-dependent baselines (i.e., $\phi(s,a)$ is a function of the state and the action).  With a state-action dependent baseline $\phi(s, a)$, the policy gradient estimator is
\begin{align}
	\hat{g}(s, a, \tau) =& \left( \hat{A}(s, a, \tau) - \phi(s, a) \right) \nabla \log \pi(a|s) \nonumber
    \\
    &+ \nabla \mathbb{E}_{a|s}\left[ \phi(s, a) \right],
    \label{eq:state-action-dependent-cv}
\end{align}
Now, $\nabla \mathbb{E}_{a|s}\left[ \phi(s, a) \right] \neq 0$ in general, so it must be analytically evaluated or estimated with low variance for the baseline to be effective. 

\begin{figure*}[h]
  \centering
  \includegraphics[width=\textwidth]{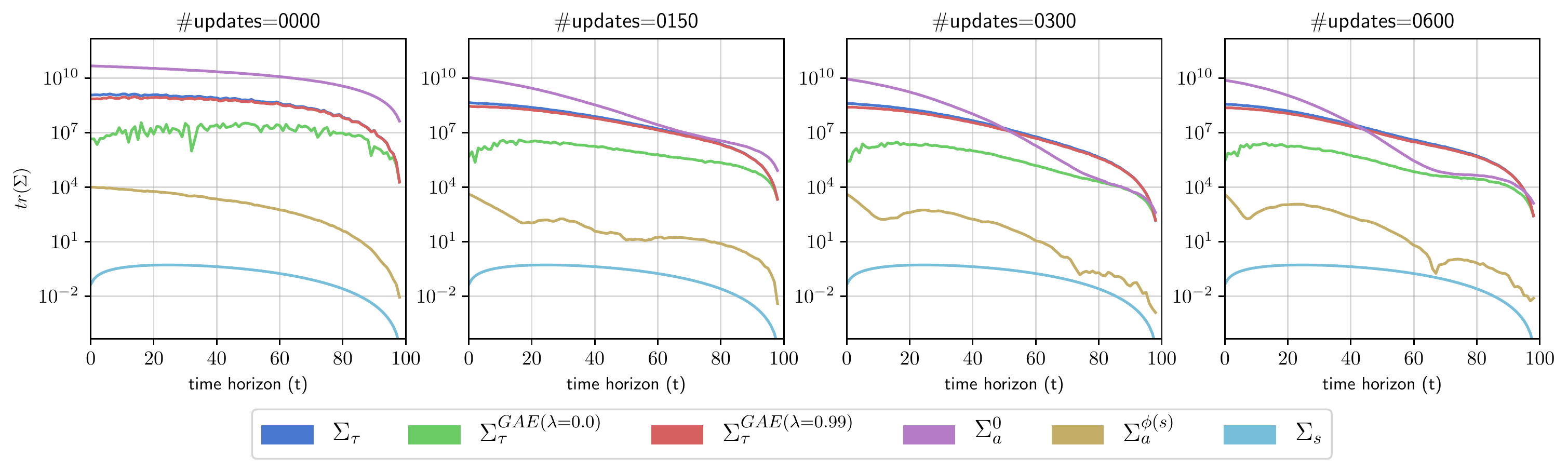}
  \caption{Evaluating the variance terms (Eq.~\ref{eq:var}) of the policy gradient estimator on a 2D point mass task (an LQG system) with finite horizon $T=100$. The total variance of the gradient estimator covariance is plotted against time in the task ($t$). Each plot from left to right corresponds to a different stage in learning (see Appendix~\ref{sec:lqr} for policy visualizations), and its title indicates the number of policy updates completed. $\Sigma_a^0$ and $\Sigma_a^{\phi(s)}$ correspond to the $\Sigma_a$ term without a baseline and using the value function as a state-dependent baseline, respectively. Importantly, an optimal state-action-dependent baseline reduces $\Sigma_a$ to $0$, so $\Sigma_a^{\phi(s)}$ upper bounds the variance reduction possible from using a state-action-dependent baseline over a state-dependent baseline. In this task, $\Sigma_a^{\phi(s)}$ is much smaller than $\Sigma_\tau$, so the reduction in overall variance from using a state-action-dependent baseline would be minimal. $\Sigma_\tau^{GAE(\lambda)}$ indicates the $\Sigma_\tau$ term with GAE-based return estimates. We include animated GIF visualizations of the variance terms and policy as learning progresses in the Supplementary Materials.
  } 
  \label{fig:lqg_var}
\end{figure*}

When the action set is discrete and not large, it is straightforward to analytically evaluate the expectation in the second term~\citep{gu2017interpolated,gruslys2017reactor}. In the continuous action case, \citet{gu2017q} set $\phi(s, a)$ to be the first order Taylor expansion of a learned advantage function approximator. Because $\phi(s, a)$ is linear in $a$, the expectation can be analytically computed. \citet{gu2017interpolated,liu2018sample,grathwohl2018backpropagation} set $\phi(s, a)$ to be a learned function approximator and leverage the reparameterization trick to estimate $\nabla \mathbb{E}_{a|s}\left[ \phi(s, a) \right]$ with low variance when $\pi$ is reparameterizable~\citep{kingma2013auto,rezende2014stochastic}.

\begin{figure*}[h]
  \centering
  \includegraphics[width=0.82\textwidth]{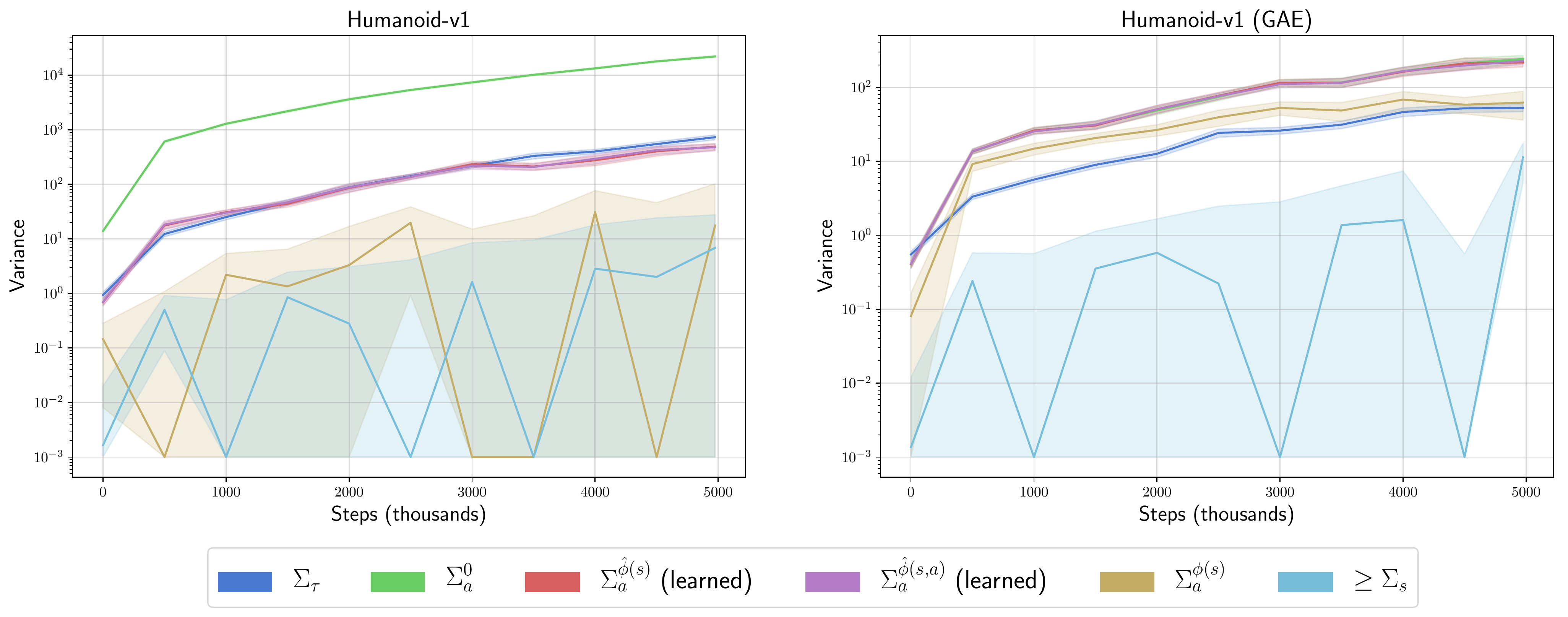}
  \caption{Evaluating the variance terms (Eq.~\ref{eq:var}) of the gradient estimator when $\hat{A}(s, a, \tau)$ is the discounted return (left) and GAE (right) with various baselines on Humanoid (See Appendix Figure~\ref{fig:halfcheetah_var} for results on HalfCheetah). The x-axis denotes the number of environment steps used for training. The policy is trained with TRPO. We set $\phi(s) = \mathbb{E}_{a|s}\left[\hat{A}(s, a)\right]$ and $\phi(s, a) = \hat{A}(s, a)$. The ``learned'' label in the legend indicates that a function approximator to $\phi$ was used instead of directly using $\phi$. Note that when using $\phi(s, a) = \hat{A}(s, a)$, $\Sigma_a^{\phi(s, a)}$ is $0$, so is not plotted. Since $\Sigma_s$ is small, we plot an upper bound on $\Sigma_s$. The upper and lower bands indicate two standard errors of the mean. In the left plot, lines for $\Sigma_a^{\hat{\phi}(s)}$ and $\Sigma_a^{\hat{\phi}(s, a)}$ overlap and in the right plot, lines for $\Sigma_a^0, \Sigma_a^{\hat{\phi}(s)}$, and $\Sigma_a^{\hat{\phi}(s, a)}$ overlap.}
  \label{fig:humanoid_var}
  \vspace{-0.1in}
\end{figure*}

\section{Policy Gradient Variance Decomposition}
\label{sec:variance_analysis}
Now, we analyze the variance of the policy gradient estimator with a state-action dependent baseline (Eq.~\ref{eq:state-action-dependent-cv}). This is an unbiased estimator of $\mathbb{E}_{s, a, \tau}\left[ \hat{A}(s, a, \tau) \nabla \log \pi(a | s) \right]$ for any choice of $\phi$. For theoretical analysis, we assume that we can analytically evaluate the expectation over $a$ in the second term because it only depends on $\phi$ and $\pi$, which we can evaluate multiple times without querying the environment.

The variance of the policy gradient estimator in Eq.~\ref{eq:state-action-dependent-cv}, $\Sigma := \Var_{s, a, \tau}(\hat{g})$, can be decomposed using the law of total variance as
\begin{equation}
\begin{split}
	%&\Sigma`' \\
    %=&~E_{s,a} \Var_{\tau|s, a} \left( \left(\hat{Q}(s, a, \tau) - \phi(s,a) \right) \nabla \log \pi(a|s) + \nabla E_{a|s} \phi(s, a) \right) \\
    %&+ \Var_{s,a} E_{\tau|s,a} \left[ \left(\hat{Q}(s, a, \tau) - \phi(s,a) \right) \nabla \log \pi(a|s) + \nabla E_{a|s} \phi(s, a) \right] \\
     %  =&~E_{s,a} \Var_{\tau|s, a} \left( \hat{Q}(s, a, \tau)\nabla \log \pi(a|s) \right) \\
    %&+ \Var_{s,a} \biggr( \left( \hat{Q}(s, a) - \phi(s,a) \right) \nabla \log \pi(a|s) + \nabla E_{a|s} \phi(s, a) \biggr), \\
    \Sigma =&~\mathbb{E}_s \left[ \Var_{a, \tau|s} \left( \left( \hat{A}(s, a, \tau) - \phi(s, a) \right) \nabla \log \pi(a|s) \right) \right] \\
    &+ \Var_s  \mathbb{E}_{a, \tau|s} \left[ \hat{A}(s, a, \tau) \nabla \log \pi(a|s) \right],
    %&+ \Var_s  \bigg( \mathbb{E}_{a, \tau|s} \left[ \left(\hat{A}(s, a, \tau) - \phi(s,a) \right) \nabla \log \pi(a|s) \right] \\
    %&+ \nabla \mathbb{E}_{a \sim \pi}\left[ \phi(s, a) \right] \bigg). 
\nonumber
\end{split}
\end{equation}
where the simplification of the second term is because the baseline does not introduce bias. We can further decompose the first term,
\begin{align*}
	%&\Var_{s,a} \biggr( \left( \hat{Q}(s, a) - \phi(s,a) \right) \nabla \log \pi(a|s) + \nabla E_{a|s} \phi(s, a) \biggr) \\
    %=& E_s \Var_{a|s}  \biggr( \left( \hat{Q}(s, a) - \phi(s,a) \right) \nabla \log \pi(a|s) \biggr) \\
    %&+ \Var_s E_{a|s} \biggr[ \hat{Q}(s, a) \nabla \log \pi(a|s) \biggr]
    \mathbb{E}_s & \left[ \Var_{a, \tau|s} \left( \left( \hat{A}(s, a, \tau) - \phi(s, a) \right) \nabla \log \pi(a|s) \right) \right] \\
	=&~ \mathbb{E}_{s, a} \left[ \Var_{\tau | s, a} \left( \hat{A}(s, a, \tau) \nabla \log \pi(a|s) \right) \right] \\
    &+ \mathbb{E}_s \left[ \Var_{a | s} \left( \left( \hat{A}(s, a)  - \phi(s, a) \right) \nabla \log \pi(a|s) \right) \right],
\end{align*}
where $\hat{A}(s, a) = \mathbb{E}_{\tau | s, a} \left[ \hat{A}(s, a, \tau) \right]$. Putting the terms together, we arrive at the following:
%\vspace{-0.2in}
%\begin{widetext}
\begin{align}
  \Sigma =& \underbrace{\mathbb{E}_{s, a} \left[ \Var_{\tau | s, a} \left( \hat{A}(s, a, \tau) \nabla \log \pi(a|s) \right) \right]}_{\Sigma_\tau} \nonumber \\
 	&+ \underbrace{\mathbb{E}_s \left[ \Var_{a | s} \left( \left( \hat{A}(s, a)  - \phi(s, a) \right) \nabla \log \pi(a|s) \right) \right]}_{\Sigma_a} \nonumber \\
    &+ \underbrace{\Var_s \left( \mathbb{E}_{a|s} \left[ \hat{A}(s, a) \nabla \log \pi(a|s) \right] \right)}_{\Sigma_s}. \label{eq:var}
\end{align}
%\end{widetext}
%\begin{align}
%\Var_{s, a, \tau} & \left( \left( \hat{A}(s, a, \tau) - \phi(s, a) %\right) \nabla \log \pi(a|s) + \nabla \mathbb{E}_{a | s} \phi(s, a) %\right) = \overbrace{\mathbb{E}_{s, a} \left[ \Var_{\tau | s, a} \left( %\hat{A}(s, a, \tau) \nabla \log \pi(a|s) \right) \right]}^{\Sigma_\tau} %\nonumber \\
% 	&+ \underbrace{\mathbb{E}_s \left[ \Var_{a | s} \left( \left( \hat{A}(s, a)  - \phi(s, a) \right) \nabla \log \pi(a|s) \right) %\right]}_{\Sigma_a} + \underbrace{\Var_s \left( \mathbb{E}_{a|s} \left[ %\hat{A}(s, a) \nabla \log \pi(a|s) \right] \right)}_{\Sigma_s}. %\label{eq:var}
%\end{align}
% \vspace{-0.2in} % SB 2/07 another option, if we don't like the lines.
% \begin{strip}
% \begin{align}
% \Sigma := \Var_{s, a, \tau} & \left( \left( \hat{A}(s, a, \tau) - \phi(s, a) \right) \nabla \log \pi(a|s) + \nabla \mathbb{E}_{a | s} \phi(s, a) \right)  = \overbrace{\mathbb{E}_{s, a} \left[ \Var_{\tau | s, a} \left( \hat{A}(s, a, \tau) \nabla \log \pi(a|s) \right) \right]}^{\Sigma_1} \nonumber \\
% 	&+ \underbrace{\mathbb{E}_s \left[ \Var_{a | s} \left( \left( \hat{A}(s, a)  - \phi(s, a) \right) \nabla \log \pi(a|s) \right) \right]}_{\Sigma_2} + \underbrace{\Var_s \left( \mathbb{E}_{a|s} \left[ \hat{A}(s, a) \nabla \log \pi(a|s) \right] \right)}_{\Sigma_3}. \label{eq:var}
% \end{align}
% \end{strip}

Notably, only $\Sigma_a$ involves $\phi$, and it is clear that the variance minimizing choice of $\phi(s, a)$ is $\hat{A}(s, a)$. For example, if $\hat{A}(s, a, \tau) = \sum_t \gamma^t r_t$, the discounted return, then the optimal choice of $\phi(s, a)$ is $\hat{A}(s,a) = \mathbb{E}_{\tau | s, a} \left[ \sum_t \gamma^t r_t \right] = Q^{\pi}(s, a)$, the state-action value function.

The variance in the on-policy gradient estimate arises from the fact that we only collect data from a limited number of states $s$, that we only take a single action $a$ in each state, and that we only rollout a single path from there on $\tau$.  Intuitively, $\Sigma_\tau$ describes the variance due to sampling a single $\tau$, $\Sigma_a$ describes the variance due to sampling a single $a$, and lastly $\Sigma_s$ describes the variance coming from visiting a limited number of states. The magnitudes of these terms depends on task specific parameters and the policy. 

The relative magnitudes of the variance terms will determine the effectiveness of the optimal state-action-dependent baseline. In particular, denoting the value of the second term when using a state-dependent baseline by $\Sigma_a^{\phi(s)}$, the variance of the policy gradient estimator with a state-dependent baseline is $\Sigma_a^{\phi(s)} + \Sigma_\tau + \Sigma_s$. When $\phi(s, a)$ is optimal, $\Sigma_a$ vanishes, so the variance is $\Sigma_\tau + \Sigma_s$. Thus, an optimal state-action-dependent baseline will be beneficial when $\Sigma_a^{\phi(s)}$ is large relative to $\Sigma_\tau + \Sigma_s$. We expect this to be the case when single actions have a large effect on the overall discounted return (e.g., in a Cliffworld domain, where a single action could cause the agent to fall of the cliff and suffer a large negative reward). Practical implementations of a state-action-dependent baseline require learning $\phi(s, a)$, which will further restrict the potential benefits.

\subsection{Variance in LQG Systems}
Linear-quadratic-Gaussian (LQG) systems~\citep{stengel1986optimal} are a family of widely studied continuous control problems with closed-form expressions for optimal controls, quadratic value functions, and Gaussian state marginals. We first analyze the variance decomposition in an LQG system because it allows nearly analytic measurement of the variance terms in Eq.~\ref{eq:var} (See Appendix~\ref{sec:lqr} for measurement details). 

Figure~\ref{fig:lqg_var} plots the variance terms for a simple 2D point mass task using discounted returns as the choice of $\hat{A}(s, a, \tau)$ (See Appendix~\ref{sec:lqr} for task details). As expected, without a baseline ($\phi = 0$), the variance of $\Sigma_a^0$ is much larger than $\Sigma_\tau$ and $\Sigma_s$. Further, using the value function as a state-dependent baseline ($\phi(s) = V^\pi(s)$), results in a large variance reduction (compare the lines for $\Sigma_a^{\phi(s)}$ and $\Sigma_a^0$ in Figure~\ref{fig:lqg_var}). An optimal state-action-dependent baseline would reduce $\Sigma_a^{\phi(s)}$ to $0$, however, for this task, such a baseline would not significantly reduce the total variance because $\Sigma_\tau$ is already large relative to $\Sigma_a^{\phi(s)}$ (Figure~\ref{fig:lqg_var}). 

We also plot the effect of using GAE\footnote{For the LQG system, we use the oracle value function to compute the GAE estimator. In the rest of the experiments, GAE is computed using a learned value function.}~\citep{schulman2015high} on $\Sigma_\tau$ for $\lambda=\{0,0.99\}$.  Baselines and GAE reduce different components of the gradient variances, and this figure compares their effects throughout the learning process.

\subsection{Empirical Variance Measurements}
We estimate the magnitude of the three terms for benchmark continuous action tasks as training proceeds. Once we decide on the form of $\phi(s, a)$, approximating $\phi$ is a learning problem in itself. To understand the approximation error, we evaluate the situation where we have access to an oracle $\phi(s, a)$ and when we learn a function approximator for $\phi(s, a)$. Estimating the terms in Eq.~\ref{eq:var} is nontrivial because the expectations and variances are not available in closed form. We construct unbiased estimators of the variance terms and repeatedly draw samples to drive down the measurement error (see Appendix~\ref{sec:est_var} for details). We train a policy using TRPO\footnote{The relative magnitudes of the variance terms depend on the task, policy, and network structures. For evaluation, we use a well-tuned implementation of TRPO (Appendix~\ref{sec:disc_app}).}~\cite{schulman2015trust} and as training proceeds, we plot each of the individual terms $\Sigma_\tau, \Sigma_a$, and $\Sigma_s$ of the gradient estimator variance for Humanoid in Figure~\ref{fig:humanoid_var} and for HalfCheetah in Appendix Figure~\ref{fig:halfcheetah_var}. Additionally, we repeat the experiment with the horizon-aware value functions (described in Section~\ref{sec:disc}) in Appendix Figures~\ref{fig:humanoid_disc_var} and~\ref{fig:halfcheetah_disc_var}. 

We plot the variance decomposition for two choices of $\hat{A}(s, a, \tau)$: the discounted return, $\sum_t \gamma^t r_t$, and GAE~\cite{schulman2015high}. In both cases, we set $\phi(s) = \mathbb{E}_{a |s}\left[\hat{A}(s, a)\right]$ and $\phi(s, a) = \hat{A}(s, a)$ (the optimal state-action-dependent baseline). When using the discounted return, we found that $\Sigma_\tau$ dominates $\Sigma_a^{\phi(s)}$, suggesting that even an optimal state-action-dependent baseline (which would reduce $\Sigma_a$ to 0) would not improve over a state-dependent baseline (Figure~\ref{fig:humanoid_var}). In contrast, with GAE, $\Sigma_\tau$ is reduced and now the optimal state-action-dependent baseline would reduce the overall variance compared to a state-dependent baseline. However, when we used function approximators to $\phi$, we found that the state-dependent and state-action-dependent function approximators produced similar variance and much higher variance than when using an oracle $\phi$ (Figure~\ref{fig:humanoid_var}). This suggests that, in practice, we would not see improved learning performance using a state-action-dependent baseline over a state-dependent baseline on these tasks. We confirm this in later experiments in Sections~\ref{sec:mirage} and \ref{sec:disc}.

Furthermore, we see that closing the function approximation gap of $V(s)$ and $\phi(s)$ would produce much larger reductions in variance than from using the optimal state-action-dependent baseline over the state-dependent baseline. This suggests that improved function approximation of both $V(s)$ and $\phi(s)$ should be a priority. Finally, $\Sigma_s$ is relatively small in both cases, suggesting that focusing on reducing variance from the first two terms of Eq.~\ref{eq:var}, $\Sigma_\tau$ and $\Sigma_a$, will be more fruitful. 
\begin{figure*}[ht] 
  \centering
  \includegraphics[width=0.95\textwidth]{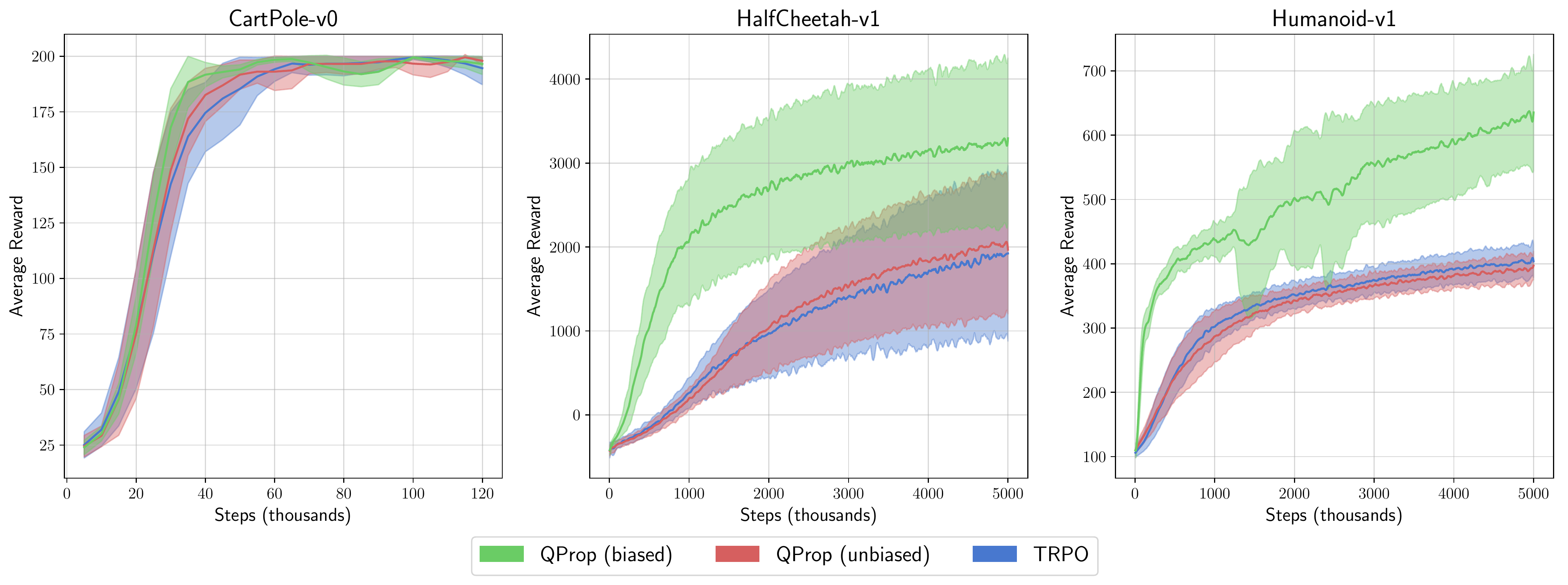}
  \caption{Evaluation of Q-Prop, an unbiased version of Q-Prop that applies the normalization to all terms, and TRPO (implementations based on the code accompanying \citet{gu2017q}). We plot mean episode reward with standard deviation intervals capped at the minimum and maximum across 10 random seeds. The batch size across all experiments was 5000. On the continuous control tasks (HalfCheetah and Humanoid), we found that that the unbiased Q-Prop performs similarly to TRPO, while the (biased) Q-Prop outperforms both. On the discrete task (CartPole), we found almost no difference between the three algorithms.}
  \label{fig:qprop}
  \vspace{-0.1in}
\end{figure*}

\section{Unveiling the Mirage} 
\label{sec:mirage}
In the previous section, we decomposed the policy gradient variance into several sources, and we found that in practice, the source of variance reduced by the state-action-dependent baseline is not reduced when a function approximator for $\phi$ is used. However, this appears to be a paradox: if the state-action-dependent baseline does not reduce variance, how are prior methods that propose state-action-dependent baselines able to report significant improvements in learning performance? We analyze implementations accompanying these works, and show that they actually introduce bias into the policy gradient due to subtle implementation decisions\footnote{The implementation of the state-action-dependent baselines for continuous control in \citep{grathwohl2018backpropagation} suffered from two critical issues (see Appendix~\ref{sec:backprop_void} for details), so it was challenging to determine the source of their observed performance. After correcting these issues in their implementation, we do not observe an improvement over a state-dependent baseline, as shown in Appendix Figure~\ref{fig:bttv}. We emphasize that these observations are restricted to the continuous control experiments as the rest of the experiments in that paper use a separate codebase that is unaffected.}. 
We find that these methods are effective not because of unbiased variance reduction, but instead because they introduce bias for variance reduction.

\subsection{Advantage Normalization}
Although Q-Prop and IPG~\citep{gu2017interpolated} (when $\nu = 0$) claim to be unbiased, the implementations of Q-Prop and IPG apply an adaptive normalization to only some of the estimator terms, which introduces a bias. Practical implementations of policy gradient methods~\citep{mnih2014neural,schulman2015high,duan2016benchmarking}
often normalize the advantage estimate $\hat{A}$, also commonly referred to as the \emph{learning signal}, to unit variance with batch statistics. This effectively serves as an adaptive learning rate heuristic that bounds the gradient variance.

The implementations of Q-Prop and IPG normalize the learning signal $\hat{A}(s, a, \tau) - \phi(s, a)$, but not the bias correction term $\nabla \mathbb{E}_{a}\left[ \phi(s, a) \right]$. Explicitly, the estimator with such a normalization is,
\begin{align*}
\hat{g}(s, a, \tau) = & \frac{1}{\hat{\sigma}} \left( \hat{A}(s, a, \tau) - \phi(s, a) - \hat{\mu} \right) \nabla \log \pi(a|s) 
    \\
    &+ \nabla \mathbb{E}_{a|s}\left[ \phi(s, a) \right],
\end{align*}
where $\hat{\mu}$ and $\hat{\sigma}$ are batch-based estimates of the mean and standard deviation of $\hat{A}(s, a, \tau) - \phi(s, a)$. This deviates from the method presented in the paper and introduces bias. In fact, IPG~\citep{gu2017interpolated} analyzes the bias in the implied objective that would be introduced when the first term has a different weight from the bias correction term, proposing such a weight as a means to trade off bias and variance. We analyze the bias and variance of the gradient estimator in Appendix~\ref{sec:app_ipg}. However, the weight actually used in the implementation is off by the factor $\hat{\sigma}$, and never one (which corresponds to the unbiased case). This introduces an adaptive bias-variance trade-off that constrains the learning signal variance to 1 by automatically adding bias if necessary. 

In Figure~\ref{fig:qprop}, we compare the implementation of Q-Prop from \citep{gu2017q}, an unbiased implementation of Q-Prop that applies the normalization to all terms, and TRPO. We found that the adaptive bias-variance trade-off induced by the asymmetric normalization is crucial for the gains observed in \citep{gu2017q}. If implemented as unbiased, it does not outperform TRPO.

\subsection{Poorly Fit Value Functions}
\label{sec:poor-value-fit}

In contrast to our results, \citet{liu2018sample} report that state-action-dependent baselines significantly reduce variance over state-dependent baselines on continuous action benchmark tasks (in some cases by six orders of magnitude). We find that this conclusion was caused by a poorly fit value function.

\begin{figure*}[!ht]
  \centering
  \includegraphics[width=0.9\textwidth]{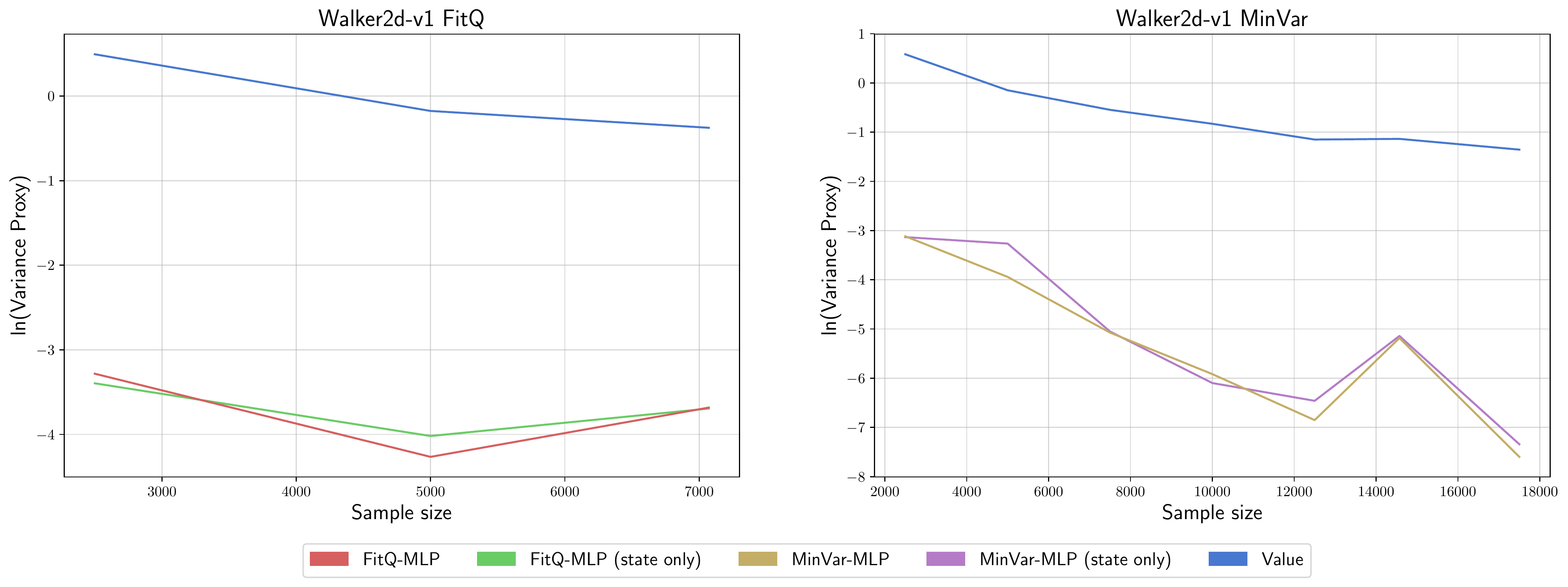}
  \caption{Evaluating an approximation of the variance of the policy gradient estimator with no additional state-dependent baseline (Value), a state-dependent baseline (state only), and a state-action-dependent baseline. FitQ indicates that the baseline was fit by regressing to Monte Carlo returns. MinVar indicates the baseline was fit by minimizing an approximation to the variance of the gradient estimator. We found that the state-dependent and state-action-dependent baselines reduce variance similarly. The large gap between Value and the rest of the methods is due to poor value function fitting. These results were generated by modifying the Stein control variate implementation~\citep{liu2018sample}.}
  \label{fig:stein_var}
\end{figure*}

The GAE advantage estimator has mean zero when $\hat{V}(s) = V^\pi(s)$, which suggests that a state-dependent baseline is unnecessary if $\hat{V}(s) \approx V^\pi(s)$. As a result, a state-dependent baseline is typically omitted when the GAE advantage estimator is used. This is the case in~\citep{liu2018sample}. However, when $\hat{V}(s)$ poorly approximates $V^\pi(s)$, the GAE advantage estimator has nonzero mean, and a state-dependent baseline can reduce variance. We show that is the case by taking the open-source code accompanying~\citep{liu2018sample}, and implementing a state-dependent baseline. It achieves comparable variance reduction to the state-action-dependent baseline (Appendix Figure~\ref{fig:stein_var}).

This situation can occur when the value function approximator is not trained sufficiently (e.g., if a small number of SGD steps are used to train $\hat{V}(s)$). Then, it can appear that adding a state-action-dependent baseline reduces variance where a state-dependent baseline would have the same effect.

\subsection{Sample-Reuse in Baseline Fitting}
Recent work on state-action-dependent baselines fits the baselines using on-policy samples~\citep{liu2018sample,grathwohl2018backpropagation} either by regressing to the Monte Carlo return or minimizing an approximation to the variance of the gradient estimator. This must be carefully implemented to avoid bias. Specifically, fitting the baseline to the current batch of data and then using the updated baseline to form the estimator results in a biased gradient~\citep{jie2010connection}.

Although this can reduce the variance of the gradient estimator, it is challenging to analyze the bias introduced. The bias is controlled by the implicit or explicit regularization (e.g., early stopping, size of the network, etc.) of the function approximator used to fit $\phi$. A powerful enough function approximator can trivially overfit the current batch of data and reduce the learning signal to $0$. This is especially important when flexible neural networks are used as the function approximators.

\citet{liu2018sample} fit the baseline using the current batch before computing the policy gradient estimator. Using the open-source code accompanying \citep{liu2018sample}, we evaluate several variants: an unbiased version that fits the state-action-dependent baseline after computing the policy step, an unbiased version that fits a state-dependent baseline after computing the policy step, and a version that estimates $\nabla E_{a|s}\left[ \phi(s, a) \right]$ with an extra sample of $a \sim \pi(a|s)$ instead of importance weighting samples from the current batch. Our results are summarized in Appendix Figure~\ref{fig:stein}. Notably, we found that using an extra sample, which should reduce variance by avoiding importance sampling, decreases performance because the baseline is overfit to the current batch. The performance of the unbiased state-dependent baseline matched the performance of the unbiased state-action-dependent baseline. On Humanoid, the biased method implemented in \citep{liu2018sample} performs best. However, on HalfCheetah, the biased methods suffer from instability.

\begin{figure*}[!ht]
  \centering
  \includegraphics[width=0.9\textwidth]{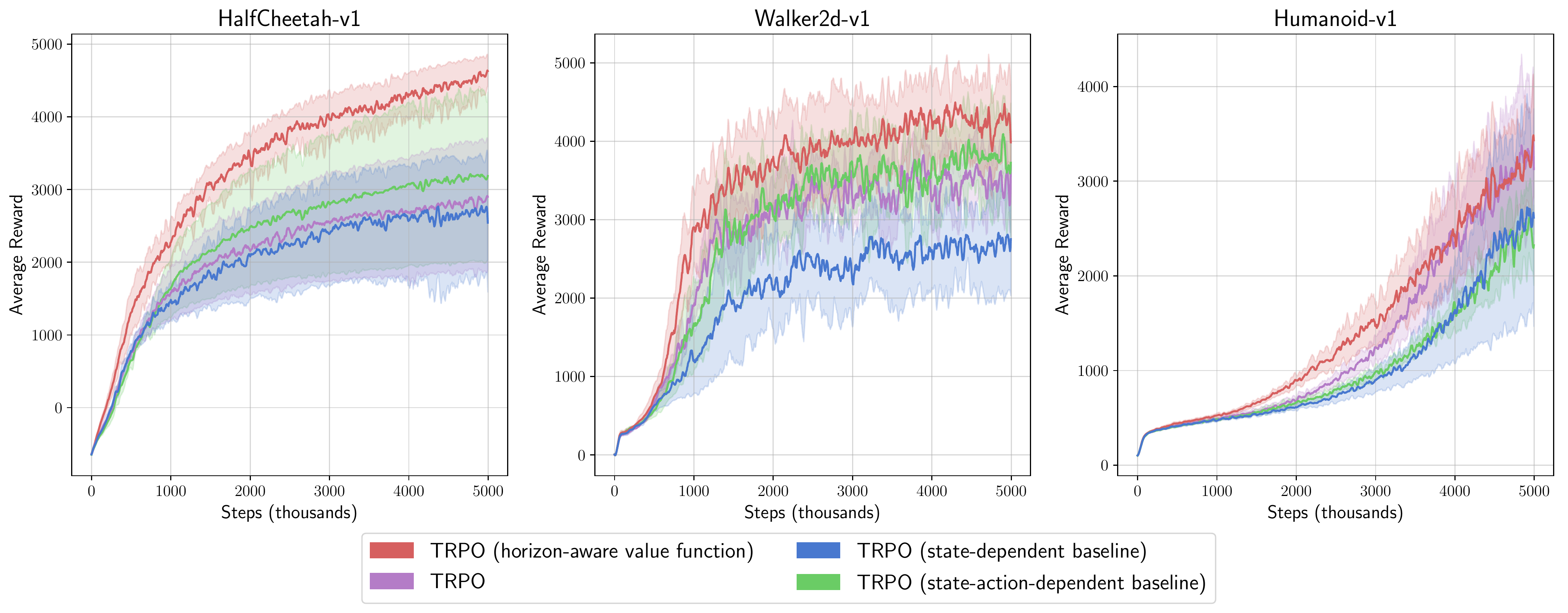}
 \caption{Evaluating the horizon-aware value function, TRPO with a state-dependent baseline, TRPO state-action-dependent baseline, and TRPO. We plot mean episode reward and standard deviation intervals capped at the minimum and maximum across 5 random seeds. The batch size across all experiments was 5000.}
  \label{fig:disc}
  \vspace{-0.2in}
\end{figure*}

\section{Horizon-Aware Value Functions}
\label{sec:disc}

\begin{figure*}[!ht]
  \centering
  \includegraphics[width=0.9\textwidth]{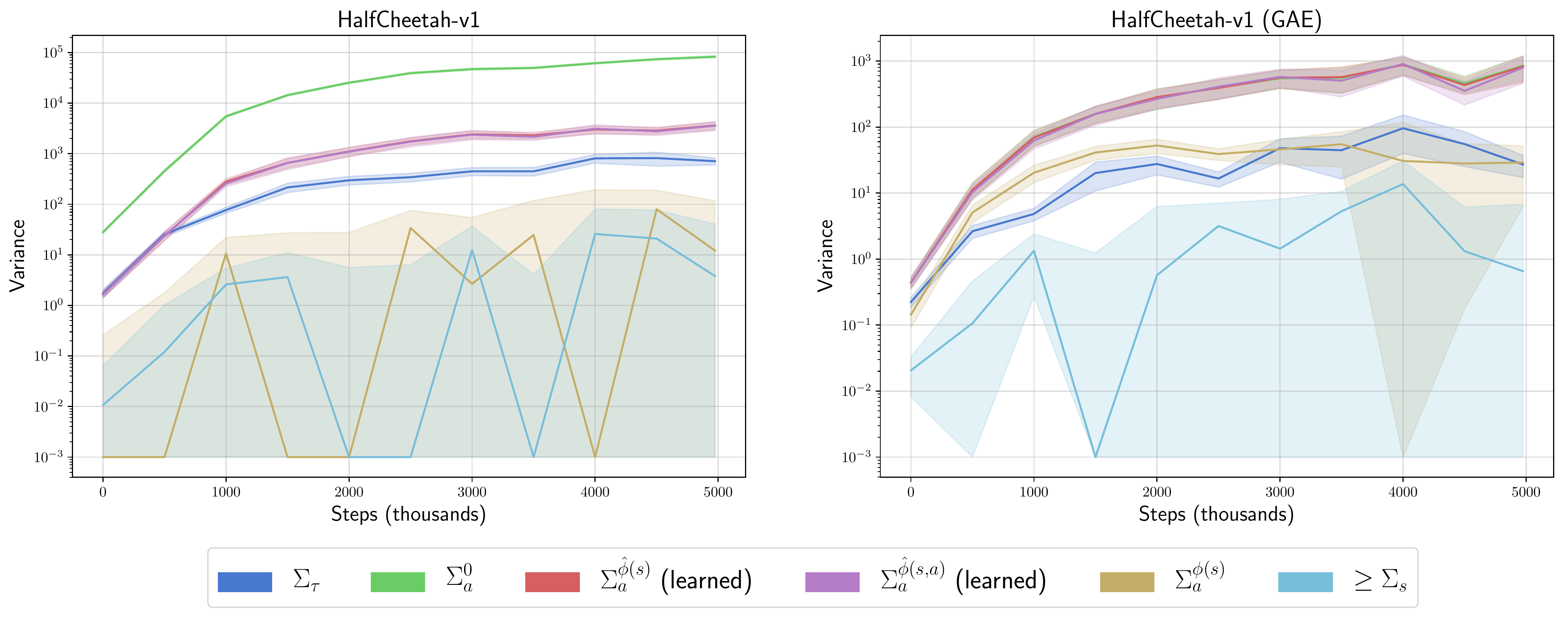}
  \caption{Evaluating the variance terms (Eq.~\ref{eq:var}) of the gradient estimator when $\hat{A}(s, a, \tau)$ is the discounted return (left) and GAE (right) with various baselines on HalfCheetah. The x-axis denotes the number of environment steps used for training. The policy is trained with TRPO. We set $\phi(s) = \mathbb{E}_{a}\left[\hat{A}(s, a)\right]$ and $\phi(s, a) = \hat{A}(s, a)$. The ``learned'' label in the legend indicates that a function approximator to $\phi$ was used instead of directly using $\phi$. Note that when using $\phi(s, a) = \hat{A}(s, a)$, $\Sigma_a^{\phi(s, a)}$ is $0$, so is not plotted. Since $\Sigma_s$ is small, we plot an upper bound on $\Sigma_s$. The upper and lower bands indicate two standard errors of the mean. In the left plot, lines for $\Sigma_a^{\hat{\phi}(s)}$ and $\Sigma_a^{\hat{\phi}(s, a)}$ overlap and in the right plot, lines for $\Sigma_a^0, \Sigma_a^{\hat{\phi}(s)}$, and $\Sigma_a^{\hat{\phi}(s, a)}$ overlap.}
  \label{fig:halfcheetah_var}
\end{figure*}

The empirical variance decomposition illustrated in Figure~\ref{fig:humanoid_var} and Appendix Figure~\ref{fig:halfcheetah_var} reveals deficiencies in the commonly used value function approximator, and as we showed in Section~\ref{sec:poor-value-fit}, a poor value approximator can produce misleading results. To fix one deficiency with the value function approximator, we propose a new horizon-aware parameterization of the value function. As with the state-action-dependent baselines, such a modification is appealing because it does not introduce bias into the underlying method.

The standard continuous control benchmarks use a fixed time horizon~\cite{duan2016benchmarking,brockman2016openai}, yet most value function parameterizations are stationary, as though the task had infinite horizon. Near the end of an episode, the expected return will necessarily be small because there are few remaining steps to accumulate reward. To remedy this, our value function approximator outputs two values: $\hat{r}(s)$ and $\hat{V}^{'}(s)$ and then we combine them with the discounted time left to form a value function estimate
\begin{align*}
\hat{V}(s_t) = \left(\sum_{i=t}^T \gamma^{i-t} \right)\hat{r}(s_t) + \hat{V}^{'}(s_t), % V_{\phi_1}(s_t) + \dfrac{1 - \gamma^{N - t}}{1 - \gamma} V_{\phi_2}(s_t)
\end{align*}
where $T$ is the maximum length of the episode. Conceptually, we can think of $\hat{r}(s)$ as predicting the average reward over future states and $\hat{V}^{'}(s)$ as a state-dependent offset. $\hat{r}(s)$ is a rate of return, so we multiply it be the remaining discounted time in the episode.
%Under simplifying assumptions, we can show that this parameterization leads to a better return estimator (Appendix~\ref{sec:app_horizon_aware}). 

\begin{figure*}
  \centering
  \includegraphics[width=0.9\textwidth]{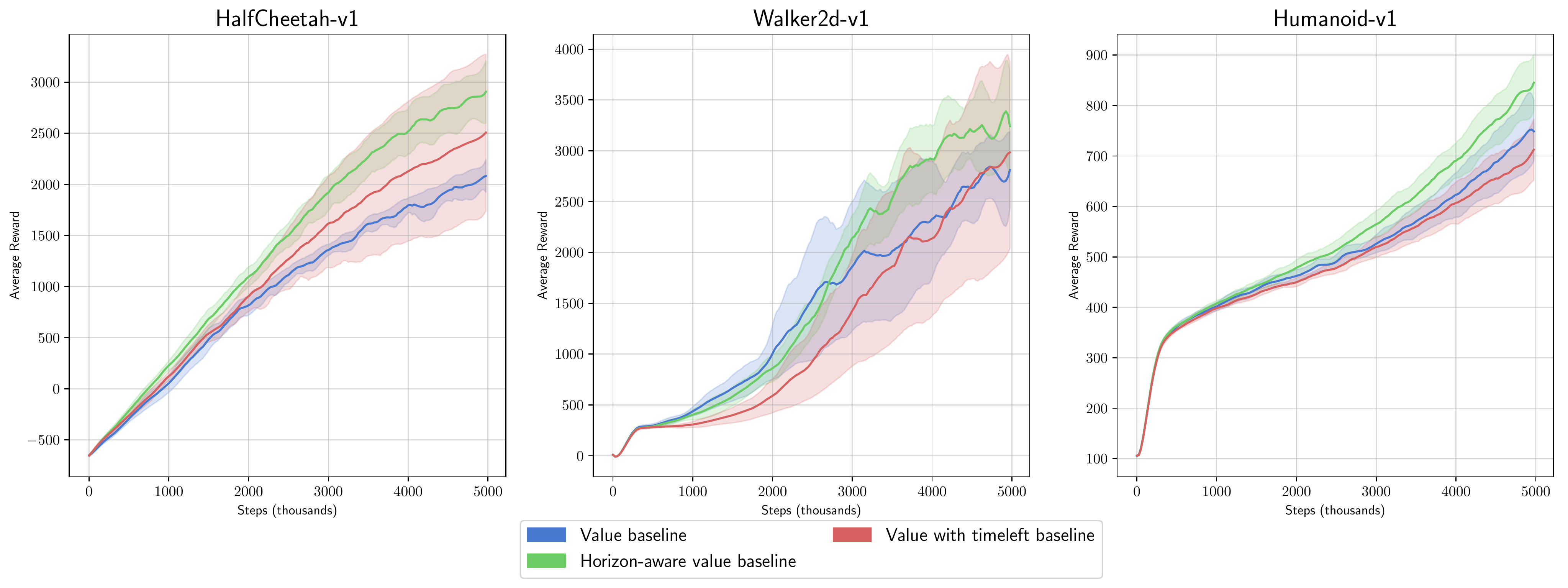}
  \caption{Evaluating the standard value function parameterization, the horizon-aware value function parameterization, and the value function with time left as an input. For these plots, we use discounted return for $\hat{A}$ and a value function baseline. We plot mean episode reward and standard deviation intervals capped at the minimum and maximum across 5 random seeds. The batch size across all experiments was 25000.}
  \label{fig:app_horizon_reward}
\end{figure*}

\begin{figure*}
  \centering
  \includegraphics[width=0.9\textwidth]{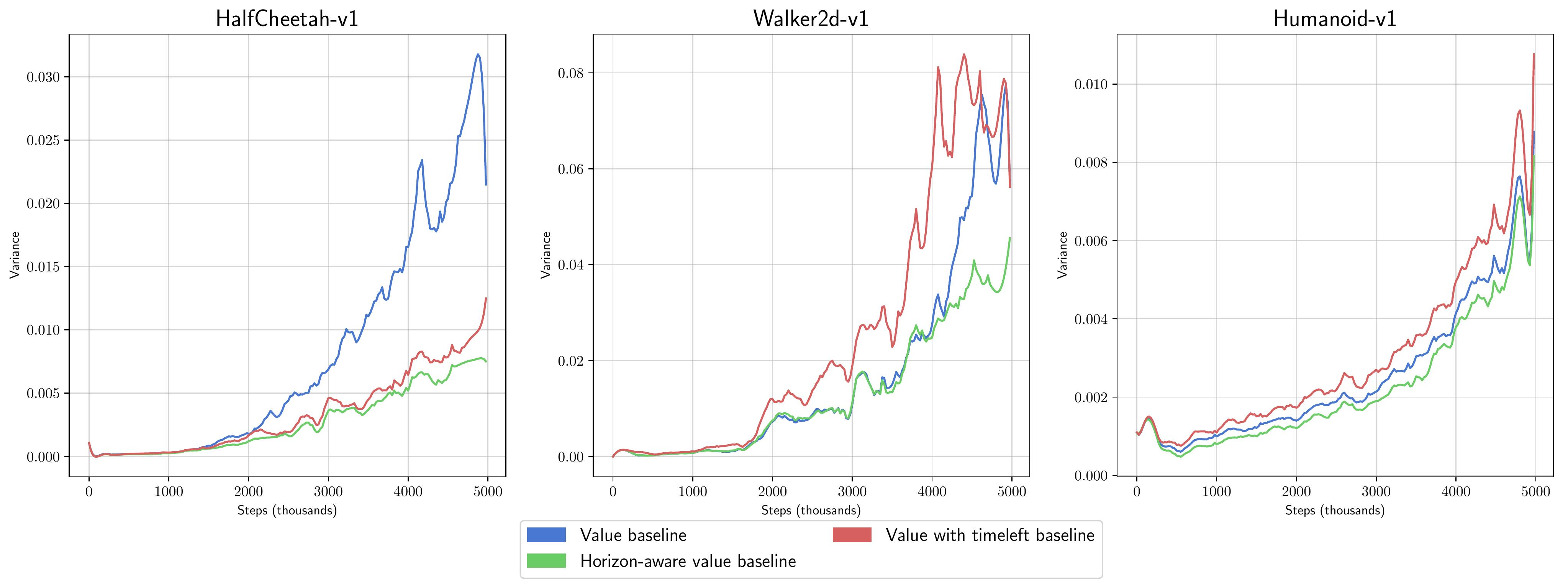}
  \caption{Evaluating the variance of the policy gradient estimator when using different value function baselines: the standard value function parameterization, the horizon-aware value function parameterization, and the value function with time left as an input. We follow the parameter trajectory from the standard value function parameterization. At each training iteration, we compute estimates of the gradient variance when using different value functions as a baseline. We average the variance estimate over all timesteps in the batch, which explains the difference in variance magnitude from previous plots. The batch size across all experiments was 25000.}
  \label{fig:app_horizon_variance}
\end{figure*}

Including time as an input to the value function can also resolve this issue (e.g., \citep{duan2016benchmarking,pardo2017time}). We compare our horizon-aware parameterization against including time as an input to the value function and find that the horizon-aware value function performs favorably (Appendix Figures~\ref{fig:app_horizon_reward} and~\ref{fig:app_horizon_variance}).

In Figure \ref{fig:disc}, we compare TRPO with a horizon-aware value function against TRPO, TRPO with a state-dependent baseline, and TRPO with a state-action-dependent baseline. Across environments, the horizon-aware value function outperforms the other methods. By prioritizing the largest variance components for reduction, we can realize practical performance improvements without introducing bias.

\section{Related Work} 
Baselines \citep{williams1992simple,weaver2001optimal} in RL fall under the umbrella of control variates, a general technique for reducing variance in Monte Carlo estimators without biasing the estimator~\citep{owen2013monte}. \citet{weaver2001optimal} analyzes the optimal state-dependent baseline, and in this work, we extend the analysis to state-action-dependent baselines in addition to analyzing the variance of the GAE estimator~\citep{tesauro1995temporal,schulman2015trust}.

\citet{dudik2011doubly} introduced the community to doubly-robust estimators, a specific form of control variate, for off-policy evaluation in bandit problems. The state-action-dependent baselines~\citep{gu2017q,wu2018variance,liu2018sample,grathwohl2018backpropagation,gruslys2017reactor} can be seen as the natural extension of the doubly robust estimator to the policy gradient setting. In fact, for the discrete action case, the policy gradient estimator with the state-action-dependent baseline can be seen as the gradient of a doubly robust estimator.
%SG.02.09: this line is fine as it is. but in my understanding doubly robust est is using (1) model as control variate, and (2) using importance weighing for first part; thus doubly robust and specific to off-policy scenarios (thus (2)). but let's leave it as it is for now...
% GT: Yes, for TRPO it's "singlely" robust, but for PPO, we do need the off-policy bit. And \phi is essentially a model.

Prior work has explored model-based~\citep{sutton1990integrated,heess2015learning,gu2016continuous} and off-policy critic-based gradient estimators~\citep{lillicrap2015continuous}. In off-policy evaluation, practitioners have long realized that constraining the estimator to be unbiased is too limiting. Instead, recent methods mix unbiased doubly-robust estimators with biased model-based estimates and minimize the mean squared error (MSE) of the combined estimator~\citep{thomas2016data,wang2016optimal}. In this direction, several recent methods have successfully mixed high-variance, unbiased on-policy gradient estimates directly with low-variance, biased off-policy or model-based gradient estimates to improve performance~\citep{o2016pgq,wang2016sample,gu2017interpolated}. It would be interesting to see if the ideas from off-policy evaluation could be further adapted to the policy gradient setting.

\section{Discussion}
State-action-dependent baselines promise variance reduction without introducing bias. In this work, we clarify the practical effect of state-action-dependent baselines in common continuous control benchmark tasks. Although an optimal state-action-dependent baseline is guaranteed not to increase variance and has the potential to reduce variance, in practice, currently used function approximators for the state-action-dependent baselines are unable to achieve significant variance reduction. Furthermore, we found that much larger gains could be achieved by instead improving the accuracy of the value function or the state-dependent baseline function approximators. 

With these insights, we re-examined previous work on state-action-dependent baselines and identified a number of pitfalls. We were also able to correctly attribute the previously observed results to implementation decisions that introduce bias in exchange for variance reduction. We intend to further explore the trade-off between bias and variance in future work.

Motivated by the gap between the value function approximator and the true value function, we propose a novel modification of the value function parameterization that makes it aware of the finite time horizon. This gave consistent improvements over TRPO, whereas the unbiased state-action-dependent baseline did not outperform TRPO. %Future work should investigate other ways to improve the value function approximation, such as off-policy fitting.
%SG.02.09: final paragraph/sentence should end with why this paper is valuable. ending with a general 'future work..., such as' without much explanation does not seem too exciting. 

Finally, we note that the relative contributions of each of the terms to the policy gradient variance are problem specific. A learned state-action-dependent baseline will be beneficial when $\Sigma_a^{\hat{\phi}(s)}$ is large relative to $\Sigma_\tau + \Sigma_s$. In this paper, we focused on continuous control benchmarks where we found this not to be the case. We speculate that in environments where single actions have a strong influence on the discounted return (and hence $\Var_a(A(s, a))$ is large), $\Sigma_a$ may be large. For example, in a discrete task with a critical decision point such as a Cliffworld domain, where a single action could cause the agent to fall of the cliff and suffer a large negative reward. Future work will investigate the variance decomposition in additional domains.

\section*{Acknowledgments} 
We thank Jascha Sohl-Dickstein, Luke Metz, Gerry Che, Yuchen Lu, and Cathy Wu for helpful discussions. We thank Hao Liu and Qiang Liu for assisting our understanding of their code. SG acknowledges support from a Cambridge-T\"{u}bingen PhD Fellowship. RET acknowledges support from Google and EPSRC grants EP/M0269571 and EP/L000776/1. ZG acknowledges support from EPSRC grant EP/J012300/1.
\bibliography{mirage}

\begin{thebibliography}{42}
\providecommand{\natexlab}[1]{#1}
\providecommand{\url}[1]{\texttt{#1}}
\expandafter\ifx\csname urlstyle\endcsname\relax
  \providecommand{\doi}[1]{doi: #1}\else
  \providecommand{\doi}{doi: \begingroup \urlstyle{rm}\Url}\fi

\bibitem[Brockman et~al.(2016)Brockman, Cheung, Pettersson, Schneider,
  Schulman, Tang, and Zaremba]{brockman2016openai}
Brockman, G., Cheung, V., Pettersson, L., Schneider, J., Schulman, J., Tang,
  J., and Zaremba, W.
\newblock Openai gym.
\newblock \emph{arXiv preprint arXiv:1606.01540}, 2016.

\bibitem[Duan et~al.(2016)Duan, Chen, Houthooft, Schulman, and
  Abbeel]{duan2016benchmarking}
Duan, Y., Chen, X., Houthooft, R., Schulman, J., and Abbeel, P.
\newblock Benchmarking deep reinforcement learning for continuous control.
\newblock In \emph{International Conference on Machine Learning}, pp.\
  1329--1338, 2016.

\bibitem[Dud{\'\i}k et~al.(2011)Dud{\'\i}k, Langford, and Li]{dudik2011doubly}
Dud{\'\i}k, M., Langford, J., and Li, L.
\newblock Doubly robust policy evaluation and learning.
\newblock \emph{arXiv preprint arXiv:1103.4601}, 2011.

\bibitem[Grathwohl et~al.(2018)Grathwohl, Choi, Wu, Roeder, and
  Duvenaud]{grathwohl2018backpropagation}
Grathwohl, W., Choi, D., Wu, Y., Roeder, G., and Duvenaud, D.
\newblock Backpropagation through the void: Optimizing control variates for
  black-box gradient estimation.
\newblock \emph{International Conference on Learning Representations (ICLR)},
  2018.

\bibitem[Gruslys et~al.(2017)Gruslys, Azar, Bellemare, and
  Munos]{gruslys2017reactor}
Gruslys, A., Azar, M.~G., Bellemare, M.~G., and Munos, R.
\newblock The reactor: A sample-efficient actor-critic architecture.
\newblock \emph{arXiv preprint arXiv:1704.04651}, 2017.

\bibitem[Gu et~al.(2016)Gu, Lillicrap, Sutskever, and Levine]{gu2016continuous}
Gu, S., Lillicrap, T., Sutskever, I., and Levine, S.
\newblock Continuous deep q-learning with model-based acceleration.
\newblock In \emph{International Conference on Machine Learning}, pp.\
  2829--2838, 2016.

\bibitem[Gu et~al.(2017{\natexlab{a}})Gu, Lillicrap, Ghahramani, Turner, and
  Levine]{gu2017q}
Gu, S., Lillicrap, T., Ghahramani, Z., Turner, R.~E., and Levine, S.
\newblock Q-prop: Sample-efficient policy gradient with an off-policy critic.
\newblock \emph{International Conference on Learning Representations (ICLR)},
  2017{\natexlab{a}}.

\bibitem[Gu et~al.(2017{\natexlab{b}})Gu, Lillicrap, Turner, Ghahramani,
  Sch{\"o}lkopf, and Levine]{gu2017interpolated}
Gu, S., Lillicrap, T., Turner, R.~E., Ghahramani, Z., Sch{\"o}lkopf, B., and
  Levine, S.
\newblock Interpolated policy gradient: Merging on-policy and off-policy
  gradient estimation for deep reinforcement learning.
\newblock In \emph{Advances in Neural Information Processing Systems}, pp.\
  3849--3858, 2017{\natexlab{b}}.

\bibitem[Heess et~al.(2015)Heess, Wayne, Silver, Lillicrap, Erez, and
  Tassa]{heess2015learning}
Heess, N., Wayne, G., Silver, D., Lillicrap, T., Erez, T., and Tassa, Y.
\newblock Learning continuous control policies by stochastic value gradients.
\newblock In \emph{Advances in Neural Information Processing Systems}, pp.\
  2944--2952, 2015.

\bibitem[Jie \& Abbeel(2010)Jie and Abbeel]{jie2010connection}
Jie, T. and Abbeel, P.
\newblock On a connection between importance sampling and the likelihood ratio
  policy gradient.
\newblock In \emph{Advances in Neural Information Processing Systems}, pp.\
  1000--1008, 2010.

\bibitem[Kakade(2002)]{kakade2002natural}
Kakade, S.~M.
\newblock A natural policy gradient.
\newblock In \emph{Advances in Neural Information Processing Systems}, pp.\
  1531--1538, 2002.

\bibitem[Kingma \& Ba(2014)Kingma and Ba]{kingma2014adam}
Kingma, D.~P. and Ba, J.
\newblock Adam: A method for stochastic optimization.
\newblock \emph{arXiv preprint arXiv:1412.6980}, 2014.

\bibitem[Kingma \& Welling(2013)Kingma and Welling]{kingma2013auto}
Kingma, D.~P. and Welling, M.
\newblock Auto-encoding variational bayes.
\newblock \emph{arXiv preprint arXiv:1312.6114}, 2013.

\bibitem[Lillicrap et~al.(2015)Lillicrap, Hunt, Pritzel, Heess, Erez, Tassa,
  Silver, and Wierstra]{lillicrap2015continuous}
Lillicrap, T.~P., Hunt, J.~J., Pritzel, A., Heess, N., Erez, T., Tassa, Y.,
  Silver, D., and Wierstra, D.
\newblock Continuous control with deep reinforcement learning.
\newblock \emph{arXiv preprint arXiv:1509.02971}, 2015.

\bibitem[Liu \& Nocedal(1989)Liu and Nocedal]{liu1989limited}
Liu, D.~C. and Nocedal, J.
\newblock On the limited memory bfgs method for large scale optimization.
\newblock \emph{Mathematical programming}, 45\penalty0 (1-3):\penalty0
  503--528, 1989.

\bibitem[Liu et~al.(2018)Liu, Feng, Mao, Zhou, Peng, and Liu]{liu2018sample}
Liu, H., Feng, Y., Mao, Y., Zhou, D., Peng, J., and Liu, Q.
\newblock Action-dependent control variates for policy optimization via stein
  identity.
\newblock \emph{International Conference on Learning Representations (ICLR)},
  2018.

\bibitem[Mnih \& Gregor(2014)Mnih and Gregor]{mnih2014neural}
Mnih, A. and Gregor, K.
\newblock Neural variational inference and learning in belief networks.
\newblock \emph{arXiv preprint arXiv:1402.0030}, 2014.

\bibitem[Mnih et~al.(2015)Mnih, Kavukcuoglu, Silver, Rusu, Veness, Bellemare,
  Graves, Riedmiller, Fidjeland, Ostrovski, et~al.]{mnih2015human}
Mnih, V., Kavukcuoglu, K., Silver, D., Rusu, A.~A., Veness, J., Bellemare,
  M.~G., Graves, A., Riedmiller, M., Fidjeland, A.~K., Ostrovski, G., et~al.
\newblock Human-level control through deep reinforcement learning.
\newblock \emph{Nature}, 518\penalty0 (7540):\penalty0 529, 2015.

\bibitem[Mnih et~al.(2016)Mnih, Badia, Mirza, Graves, Lillicrap, Harley,
  Silver, and Kavukcuoglu]{mnih2016asynchronous}
Mnih, V., Badia, A.~P., Mirza, M., Graves, A., Lillicrap, T., Harley, T.,
  Silver, D., and Kavukcuoglu, K.
\newblock Asynchronous methods for deep reinforcement learning.
\newblock In \emph{International Conference on Machine Learning}, pp.\
  1928--1937, 2016.

\bibitem[O'Donoghue et~al.(2016)O'Donoghue, Munos, Kavukcuoglu, and
  Mnih]{o2016pgq}
O'Donoghue, B., Munos, R., Kavukcuoglu, K., and Mnih, V.
\newblock Pgq: Combining policy gradient and q-learning.
\newblock \emph{arXiv preprint arXiv:1611.01626}, 2016.

\bibitem[Owen(2013)]{owen2013monte}
Owen, A.~B.
\newblock Monte carlo theory, methods and examples.
\newblock \emph{Monte Carlo Theory, Methods and Examples. Art Owen}, 2013.

\bibitem[Pardo et~al.(2017)Pardo, Tavakoli, Levdik, and
  Kormushev]{pardo2017time}
Pardo, F., Tavakoli, A., Levdik, V., and Kormushev, P.
\newblock Time limits in reinforcement learning.
\newblock \emph{arXiv preprint arXiv:1712.00378}, 2017.

\bibitem[Peters \& Schaal(2006)Peters and Schaal]{peters2006policy}
Peters, J. and Schaal, S.
\newblock Policy gradient methods for robotics.
\newblock In \emph{Intelligent Robots and Systems, 2006 IEEE/RSJ International
  Conference on}, pp.\  2219--2225. IEEE, 2006.

\bibitem[Rezende et~al.(2014)Rezende, Mohamed, and
  Wierstra]{rezende2014stochastic}
Rezende, D.~J., Mohamed, S., and Wierstra, D.
\newblock Stochastic backpropagation and approximate inference in deep
  generative models.
\newblock \emph{arXiv preprint arXiv:1401.4082}, 2014.

\bibitem[Schulman et~al.(2015{\natexlab{a}})Schulman, Levine, Abbeel, Jordan,
  and Moritz]{schulman2015trust}
Schulman, J., Levine, S., Abbeel, P., Jordan, M., and Moritz, P.
\newblock Trust region policy optimization.
\newblock In \emph{International Conference on Machine Learning}, pp.\
  1889--1897, 2015{\natexlab{a}}.

\bibitem[Schulman et~al.(2015{\natexlab{b}})Schulman, Moritz, Levine, Jordan,
  and Abbeel]{schulman2015high}
Schulman, J., Moritz, P., Levine, S., Jordan, M., and Abbeel, P.
\newblock High-dimensional continuous control using generalized advantage
  estimation.
\newblock \emph{arXiv preprint arXiv:1506.02438}, 2015{\natexlab{b}}.

\bibitem[Schulman et~al.(2017)Schulman, Wolski, Dhariwal, Radford, and
  Klimov]{schulman2017proximal}
Schulman, J., Wolski, F., Dhariwal, P., Radford, A., and Klimov, O.
\newblock Proximal policy optimization algorithms.
\newblock \emph{arXiv preprint arXiv:1707.06347}, 2017.

\bibitem[Silver et~al.(2014)Silver, Lever, Heess, Degris, Wierstra, and
  Riedmiller]{silver2014deterministic}
Silver, D., Lever, G., Heess, N., Degris, T., Wierstra, D., and Riedmiller, M.
\newblock Deterministic policy gradient algorithms.
\newblock In \emph{ICML}, 2014.

\bibitem[Silver et~al.(2016)Silver, Huang, Maddison, Guez, Sifre, Van
  Den~Driessche, Schrittwieser, Antonoglou, Panneershelvam, Lanctot,
  et~al.]{silver2016mastering}
Silver, D., Huang, A., Maddison, C.~J., Guez, A., Sifre, L., Van Den~Driessche,
  G., Schrittwieser, J., Antonoglou, I., Panneershelvam, V., Lanctot, M.,
  et~al.
\newblock Mastering the game of go with deep neural networks and tree search.
\newblock \emph{nature}, 529\penalty0 (7587):\penalty0 484--489, 2016.

\bibitem[Stengel(1986)]{stengel1986optimal}
Stengel, R.~F.
\newblock \emph{Optimal control and estimation}.
\newblock Courier Corporation, 1986.

\bibitem[Sutton(1990)]{sutton1990integrated}
Sutton, R.~S.
\newblock Integrated architectures for learning, planning, and reacting based
  on approximating dynamic programming.
\newblock In \emph{Machine Learning Proceedings 1990}, pp.\  216--224.
  Elsevier, 1990.

\bibitem[Sutton \& Barto(1998)Sutton and Barto]{sutton1998reinforcement}
Sutton, R.~S. and Barto, A.~G.
\newblock \emph{Reinforcement Learning: An Introduction}, volume~1.
\newblock MIT Press Cambridge, 1998.

\bibitem[Sutton et~al.(2000)Sutton, McAllester, Singh, and
  Mansour]{sutton2000policy}
Sutton, R.~S., McAllester, D.~A., Singh, S.~P., and Mansour, Y.
\newblock Policy gradient methods for reinforcement learning with function
  approximation.
\newblock In \emph{Advances in Neural Information Processing Systems}, pp.\
  1057--1063, 2000.

\bibitem[Tesauro(1995)]{tesauro1995temporal}
Tesauro, G.
\newblock Temporal difference learning and td-gammon.
\newblock \emph{Communications of the ACM}, 38\penalty0 (3):\penalty0 58--68,
  1995.

\bibitem[Thomas(2014)]{thomas2014bias}
Thomas, P.
\newblock Bias in natural actor-critic algorithms.
\newblock In \emph{International Conference on Machine Learning}, pp.\
  441--448, 2014.

\bibitem[Thomas \& Brunskill(2016)Thomas and Brunskill]{thomas2016data}
Thomas, P. and Brunskill, E.
\newblock Data-efficient off-policy policy evaluation for reinforcement
  learning.
\newblock In \emph{International Conference on Machine Learning}, pp.\
  2139--2148, 2016.

\bibitem[Thomas \& Brunskill(2017)Thomas and Brunskill]{thomas2017policy}
Thomas, P.~S. and Brunskill, E.
\newblock Policy gradient methods for reinforcement learning with function
  approximation and action-dependent baselines.
\newblock \emph{arXiv preprint arXiv:1706.06643}, 2017.

\bibitem[Wang et~al.(2016{\natexlab{a}})Wang, Agarwal, and
  Dudik]{wang2016optimal}
Wang, Y.-X., Agarwal, A., and Dudik, M.
\newblock Optimal and adaptive off-policy evaluation in contextual bandits.
\newblock \emph{arXiv preprint arXiv:1612.01205}, 2016{\natexlab{a}}.

\bibitem[Wang et~al.(2016{\natexlab{b}})Wang, Bapst, Heess, Mnih, Munos,
  Kavukcuoglu, and de~Freitas]{wang2016sample}
Wang, Z., Bapst, V., Heess, N., Mnih, V., Munos, R., Kavukcuoglu, K., and
  de~Freitas, N.
\newblock Sample efficient actor-critic with experience replay.
\newblock \emph{arXiv preprint arXiv:1611.01224}, 2016{\natexlab{b}}.

\bibitem[Weaver \& Tao(2001)Weaver and Tao]{weaver2001optimal}
Weaver, L. and Tao, N.
\newblock The optimal reward baseline for gradient-based reinforcement
  learning.
\newblock In \emph{Proceedings of the Seventeenth conference on Uncertainty in
  artificial intelligence}, pp.\  538--545. Morgan Kaufmann Publishers Inc.,
  2001.

\bibitem[Williams(1992)]{williams1992simple}
Williams, R.~J.
\newblock Simple statistical gradient-following algorithms for connectionist
  reinforcement learning.
\newblock In \emph{Reinforcement Learning}, pp.\  5--32. Springer, 1992.

\bibitem[Wu et~al.(2018)Wu, Rajeswaran, Duan, Kumar, Bayen, Kakade, Mordatch,
  and Abbeel]{wu2018variance}
Wu, C., Rajeswaran, A., Duan, Y., Kumar, V., Bayen, A.~M., Kakade, S.,
  Mordatch, I., and Abbeel, P.
\newblock Variance reduction for policy gradient with action-dependent
  factorized baselines.
\newblock \emph{International Conference on Learning Representations (ICLR)},
  2018.

\end{thebibliography}
\bibliographystyle{icml2018}

\section*{Appendix}

\section{Experiment Details}\label{sec:experimental_details}

\subsection{Q-Prop Experiments}
We modified the Q-Prop implementation published by the authors at \url{https://github.com/shaneshixiang/rllabplusplus} (commit: 4d55f96). We used the conservative variant of Q-Prop, as is used throughout the experimental section in the original paper. We used the default choices of policy and value functions, learning rates, and other hyperparameters as dictated by the code. 

We used a batch size of 5000 steps. We ran each of the three algorithms on a discrete (CartPole-v0) and two continuous (HalfCheetah-v1, Humanoid-v1) environments environments using OpenAI Gym~\cite{brockman2016openai} and Mujoco 1.3.

To generate the TRPO and (biased) Q-Prop results, we run the code as is.  For the unbiased Q-Prop, recall the expression for the biased gradient estimator:
\begin{align*}
\hat{g}(s, a, \tau) = & \frac{1}{\hat{\sigma}} \left( \hat{A}(s, a, \tau) - \phi(s, a) - \hat{\mu} \right) \nabla \log \pi(a|s) 
    \\
    &+ \nabla \mathbb{E}_{a}\left[ \phi(s, a) \right],
\end{align*}
To debias the Q-Prop gradient estimator, we divide the bias correction term, $\nabla \mathbb{E}_{a}\left[ \phi(s, a) \right]$, by $\hat{\sigma}$. 

\subsection{Stein Control Variate Experiments}
We used the Stein control variate implementation published by the authors at \url{https://github.com/DartML/PPO-Stein-Control-Variate} (commit: 6eec471). We used the default hyperparameters and test on two continuous control environments, HalfCheetah-v1 and Humanoid-v1 using OpenAI Gym and Mujoco 1.3.

We evaluated five algorithms in this experiment: PPO, the Stein control variates algorithm as implemented by \citep{liu2018sample}, a variant of the biased Stein algorithm that does not use importance sampling to compute the bias correction term (described below), an unbiased state-dependent baseline, and an unbiased state-action-dependent Stein baseline. All of the learned baselines were trained to minimize the approximation to the variance of the gradient estimator as described in \citep{liu2018sample}.

We use the code as is to run the first two variants. In the next variant, we estimate $\nabla E_{a \sim \pi}\left[ \phi(s, a) \right]$ with an extra sample of $a \sim \pi(a|s)$ instead of importance weighting samples from the current batch (see Eq. 20 in \citep{liu2018sample}. For the unbiased baselines, we ensure that the policy update steps for the current batch are performed before updating the baselines.

\subsection{Backpropagating through the Void}\label{sec:backprop_void}
We used the implementation published by the authors (\url{https://github.com/wgrathwohl/BackpropThroughTheVoidRL}, commit: 0e6623d) with the following modification: we measure the variance of the policy gradient estimator. In the original code, the authors accidentally measure the variance of a gradient estimator that neither method uses. We note that \citet{grathwohl2018backpropagation} recently corrected a bug in the code that caused the LAX method to use a different advantage estimator than the base method. We use this bug fix.

\subsection{Horizon-Aware Value Function Experiments}
\label{sec:disc_app}
For these experiments, we modify the open-source TRPO implementation: \url{https://github.com/ikostrikov/pytorch-trpo} (commit: 27400b8). We test four different algorithms on three different continuous control algorithms: HalfCheetah-v1, Walker2d-v1, and Humanoid-v1 using OpenAI Gym and Mujoco 1.3. 

The policy network is a two-layer MLP with 64 units per hidden layer with tanh nonlinearities. It parameterizes the mean and standard deviation of a Gaussian distribution from which actions are sampled. The value function is parameterized similarly. We implement the two outputs of the horizon-aware value function as a single neural network with two heads. Both heads have a separate linear layer off of the last hidden layer. We estimate the value functions by regressing on Monte Carlo returns. We optimize the value functions with L-BFGS~\cite{liu1989limited}. We use $\gamma = 0.99$ and $\lambda = 0.95$ for GAE, a batch size of 5000 steps, and the maximum time per episode is set to 1000. 

For the experiments in which we train an additional state-dependent or state-action-dependent baseline on the advantage estimates, we parameterize those similarly to the normal value function and train them with a learning rate of $1\mathrm{e}{-3}$ optimized with Adam~\cite{kingma2014adam}. We fit the baselines to minimize the mean squared error of predicting the GAE estimates. With the state-action-dependent baseline, we estimate $\nabla \mathbb{E}_{\pi(a|s)}\left[ \phi(s, a) \right]$ using the reparameterization trick as in \citep{liu2018sample,grathwohl2018backpropagation}.

\section{Variance computations in Linear-Quadratic-Gaussian (LQG) systems}
\label{sec:lqr}
Linear-Quadratic-Gaussian (LQG) systems~\citep{stengel1986optimal} are one of the most studied control/continuous states and actions RL problems. The LQG problems are expressed in the following generic form for a finite horizon scenario. To simplify exposition, we focus on open-loop control without observation matrices, however, it is straightforward to extend our analysis.
\begin{align*}
&p(s_0) =  \mathcal{N}(\mu^S_0, \Sigma^S_0) \\
%&\pi(a_t | s_t) = \mathcal{N}(\mu^{A|S}_t, \Sigma^{A|S}_t), \mu^{A|S}_t = K_t s_t \\
&\pi(a_t ) = \mathcal{N}(\mu^{A}_t, \Sigma^{A}_t)\\
&p(s_{t+1} | s_{t}, a_{t}) = \mathcal{N}(\mu^{S|SA}_{t+1}, \Sigma^{S|SA}_{t+1}), \mu^{S|SA}_{t+1} = A_t s_t + B_t a_t \\
&J(\theta) = E_{s_{0:T},a_{0:T}}\left[ \sum_{t=0}^T \gamma^t r_t \right],r_{t} = - s_t^T Q_t s_t - a_t^T R_t a_t
\end{align*}
The policy parameters, dynamics parameters, and reward parameters are 
%$\theta=\{K_{0:T},\Sigma_{0:T}^{A|S}\}$,
$\theta=\{\mu_{0:T}^{A},\Sigma_{0:T}^{A}\},$
$\Phi=\{\mu^S_0, \Sigma^S_0, A_{0:T-1}, B_{0:T-1}, \Sigma^{S|SA}_{1:T}\}$, and $\Psi=\{Q_{0:T},R_{0:T}\}$, respectively. While the goal of LQG is to find control parameters $\theta^*=\arg\max_{\theta} J(\theta)$, which can be solved analytically through dynamic programming, we are interested in analyzing its gradient estimator properties when a policy gradient algorithm is applied. 

\subsection{Computing $Q(s,a), V(s), A(s,a)$}
In a LQG system, both the state-action value function $Q(s_t,a_t)$ and state value function $V(s_t)$ corresponding to policy $\theta$ can be derived analytically. To achieve that, we first note that the marginal state distributions $p(s_{t+k})$ at $k\geq 1$ given \mbox{$p(s_{t}) = \mathcal{N}(\mu_{t}^S, \Sigma^S_{t})$} can be computed iteratively by the following equation,
\begin{align*}
%& p(s_{t-1},a_{t-1})=\mathcal{N}(\mu_{t-1}^{SA}, \Sigma^{SA}_{t-1}) \\
%& \quad\mu_{t-1}^{SA} = \begin{bmatrix}
%    \mu_{t-1}^S     \\
%    K_{t-1}\mu_{t-1}^S      
%\end{bmatrix},\\
%&\quad\Sigma^{SA}_{t-1}=
%\begin{bmatrix}
%   \Sigma^S_{t-1}       & \Sigma^S_{t-1} K_{t-1}^T   \\
%   K_{t-1}\Sigma^S_{t-1}       & K_{t-1} \Sigma^S_{t-1} K_{t-1}^T + \Sigma^{A|S}_{t-1}
%\end{bmatrix} \\
%& p(s_{t+1}) = \mathcal{N}(\mu_{t+1}^S, \Sigma^S_{t+1}) \\
%& \quad \mu_{t}^S = (A_t  +  K_{t-1}  )\mu_{t-1}^S ,\\
%& \quad \mu_{t+1}^S = A_t \mu_{t}^S + B_t \mu_{t}^{A},\\
%&\quad \Sigma_{t+1}^S = A_t \Sigma_{t}^S A^T_t + B_t \Sigma_{t}^A B^T_t +\Sigma_{t+1}^{S|SA}\\
& p(s_{t+k}) = \mathcal{N}(\mu_{t+k}^S, \Sigma^S_{t+k}) \\
&\quad \mu_{t+k}^S = L_{t,k} A_t\mu_{t}^S +m_{t,k} \\
&\quad \Sigma_{t+k}^S = L_{t,k} A_t \Sigma_{t}^S \left(L_{t,k} A_t\right)^T + M_{t,k}\\
&\qquad L_{t,k} = (\prod_{i=1}^{k-1} A_{t+i}^T)^T = A_{t+k-1} L_{t,k-1}\\
%=A_{t+k-1}A_{t+k-2}\dots A_{t+1} \\
&\qquad m_{t,k} = \sum_{j=0}^{k-1} L_{t+j,k-j} B_{t+j} \mu^A_{t+j} \\
&\qquad \qquad = A_{t+k-1} m_{t,k-1} + B_{t+k-1}\mu_{t+k-1}^A \\
&\qquad M_{t,k} = \sum_{j=0}^{k-1} L_{t+j,k-j} (B_{t+j} \Sigma^A_{t+j}B_{t+j}^T + \Sigma^{S|SA}_{t+j+1})L_{t+j,k-j}^T \\
&\qquad \qquad = A_{t+k-1} M_{t,k-1}  A_{t+k-1}^T \\
&\qquad \qquad + B_{t+k-1} \Sigma^A_{t+k-1}B_{t+k-1}^T + \Sigma^{S|SA}_{t+k} \\
&\qquad   L_{t,1}=I, m_{t,0}=0, M_{t,0}=0
%& \quad \Sigma_t^S = A_t \Sigma_{t-1}^S A^T_t + B_t K_{t-1} \Sigma_{t-1}^S A^T_t +\\
%&\qquad \qquad A_t \Sigma_{t-1}^S K_{t-1}^T B_t^T + B_t K_{t-1}\Sigma_{t-1}^S K_{t-1}^T B_t^T + \\
%& \qquad \qquad B_t\Sigma_{t-1}^{A|S}B^T_t + \Sigma_{t}^{S|SA}
\end{align*}
To compute $Q(s_t,a_t)$, we simply modify the above to first compute all future state-action-conditional marginals $p(s_{t+k}|s_t, a_t)$,
\begin{align*}
& p(s_{t+k}|s_t,a_t) = \mathcal{N}(\mu_{t+k}^{S|s_t,a_t}, \Sigma^{S|s_t,a_t}_{t+k}) \\
&\quad \mu_{t+k}^{S|s_t,a_t} = L_{t,k} A_t s_t + L_{t,k} B_t a_t + m_{t+1,k-1} \\
&\quad \Sigma_{t+k}^{S|s_t,a_t} = L_{t,k} \Sigma_{t+1}^{S|SA} L_{t,k}^T + M_{t+1,k-1}\\
\end{align*}
and integrate with respect to quadratic costs at each time step,
\begin{align*}
&Q(s_t,a_t) =  r_t + \sum_{k=1}^{T-t}\gamma^k E_{p(s_{t+k}|s_t,a_t)\pi(a_{t+k})}[r_{t+k}]\\
&= - \sum_{k=1}^{T-t} \gamma^k\biggl(\left(\mu_{t+k}^{S|s_t,a_t}\right)^T Q_{t+k} \mu_{t+k}^{S|s_t,a_t}\\
&\quad +\text{Tr}\left(Q_{t+k}\Sigma_{t+k}^{S|s_t,a_t}\right)\\
&\quad + (\mu_{t+k}^{A})^T R_{t+k} \mu_{t+k}^{A} + \text{Tr}\left(R_{t+k}\Sigma_{t+k}^{A}\right) \biggr)+r_t \\
&= - \biggl( s_t^T P^{SS}_{t} s_t + a_t^T P^{AA}_{t} a_t+ s_t^T P^{SA}_{t} a_t \\
&\quad + s^T_t p^{S}_{t} + a^T_t p^{A}_{t} + c_t \biggr), \\
\end{align*}
whose resulting coefficients are,
\begin{align*}
&\quad P_t^{SS} = Q_t + \sum_{k=1}^{T-t} \gamma^k \left(L_{t,k}A_t\right)^T Q_{t+k} L_{t,k}A_t \\
&\quad P_t^{AA} = R_t + \sum_{k=1}^{T-t} \gamma^k \left(L_{t,k}B_t\right)^T Q_{t+k} L_{t,k}B_t \\
&\quad P_t^{SA} = 2\sum_{k=1}^{T-t} \gamma^k \left(L_{t,k}A_t\right)^T Q_{t+k} L_{t,k}B_t \\
&\quad p_t^{S} = 2\sum_{k=1}^{T-t} \gamma^k \left(L_{t,k}A_t\right)^T Q_{t+k} m_{t+1,k-1} \\
&\quad p_t^{A} = 2\sum_{k=1}^{T-t} \gamma^k \left(L_{t,k}B_t\right)^T Q_{t+k} m_{t+1,k-1},
\end{align*}
and $c_t$ denotes constant terms that do not depend on $s_t$ or $a_t$. Note that the $Q(s_t,a_t)$ is non-stationary due to the finite time horizon. Given the quadratic $Q(s_t,a_t)$, it is straight-forward to derive the value function $V(s_t)$,
\begin{align*}
&V(s_t) = - \biggl( s_t^T P^{SS}_{t} s_t + \left(\mu_t^A\right)^T P^{AA}_{t} \mu^A_t+ s_t^T P^{SA}_{t} \mu^A_t \\
&\quad + s^T_t p^{S}_{t} + \left(\mu^A_t\right)^T p^{A}_{t} + \text{Tr}\left(P_t^{AA}\Sigma_t^A \right) + c_t \biggr), \\
\end{align*}
and the advantage function $A(s_t,a_t)$,
\begin{align*}
&A(s_t,a_t) = -\biggl( a_t^T P^{AA}_{t} a_t+ s_t^T P^{SA}_{t} a_t + a^T_t p^{A}_{t} \\
&\quad - \left(\mu_t^A\right)^T P^{AA}_{t} \mu^A_t- s_t^T P^{SA}_{t} \mu^A_t - \left(\mu^A_t\right)^T p^{A}_{t}\\
&\quad - \text{Tr}\left(P_t^{AA}\Sigma_t^A \right) \biggr) \\
&= -\biggl( a_t^T P^{AA}_{t} a_t+ s_t^T P^{SA}_{t} a_t + s^T_t \tilde{p}^{S}_{t} + a^T_t p^{A}_{t} + \tilde{c}_t \biggr) \\
&\quad \tilde{p}^{S}_{t} = P_t^{SA} \mu_{t}^A, \\
&\quad \tilde{c}_t = - \left(\mu_t^A\right)^T P^{AA}_{t} \mu^A_t -  \left(\mu^A_t\right)^T p^{A}_{t} - \text{Tr}\left(P_t^{AA}\Sigma_t^A \right). 
\end{align*}

\subsection{Computing the LQG analytic gradients}
Given quadratic $Q(s_t,a_t)$ and $A(s_t,a_t)$, it is tractable to compute the exact gradient with respect to the Gaussian parameters. For the gradient with respect to the mean $\mu_t^A$, the derivation is given below, where we drop the time index $t$ since it applies to every time step.
\begin{align*}
g^\mu(s) &= E_{a} [Q(s,a) \nabla_{\mu} \log \pi(a) ] \\
&= -E_a [((s^T P^{SA} a + a^T P^{AA} a + a^T_t p_t^A) \\
&\quad \cdot (a^T - \mu^T) \left(\Sigma^A\right)^{-1})^T] \\
&=-\biggl( (P^{SA})^T s + 2 P^{AA} \mu + p_t^A \biggr)
\end{align*}
Written with time indices, the state-conditional gradient and the full gradient are tractably expressed as the following for both $Q(s_t,a_t)$ and $A(s_t,a_t)$-based estimators,
\begin{align*}
&g^{\mu}(s_t) = -\biggl((P^{SA}_t)^Ts_t + 2 P^{AA}_t \mu^A_t + p_t^A \biggr)\\
&g^{\mu}_{t} = -\biggl((P_t^{SA})^T\mu_t^S + 2 P^{AA} \mu^A_t + p_t^A \biggr).
\end{align*}
Similarly, the gradients with respect to the covariance $\Sigma_t^A$ can be computed analytically. For the LQG system, we analyze the variance with respect to the mean parameters, so we omit the derivations here.

\subsection{Estimating the LQG variances}
Given analytic expressions for $Q(s_t,a_t)$, $A(s_t,a_t)$ and $g(s_t)$, it is simple to estimate the variance terms $\Sigma_\tau$, $\Sigma_a$, and $\Sigma_s$ in Eq.~\ref{eq:var}. $\Sigma_s$ is the simplest and given by,
\begin{align*}
\Sigma_{s,t}^\mu &= \text{Var}_{s_t} E_{a_t} \biggl[ Q(s_t,a_t) \nabla_{\mu} \log \pi(a_t) \biggr] \\
&= \text{Var}_{s_t} \biggl(- (P^{SA}_t)^Ts_t - 2 P^{AA}_t \mu^A_t - p_t^A \biggr) \\
&= (P^{SA}_t)^T \Sigma_{t}^S P^{SA}_t.
\end{align*}
$\Sigma_a$ term can also be computed analytically; however, to avoid integrating complex fourth-order terms, we use a sample-based estimate over $s$ and $a$, which is sufficiently low variance. For the gradient estimator with $Q(s_t,a_t)$, the variance expression is,
\begin{align*}
\Sigma_{a,t}^{\mu,Q} &= E_{s_t} \text{Var}_{a_t}\biggl( Q(s_t,a_t) \nabla_{\mu} \log \pi(a_t)  \biggr) \\
&= E_{s_t} \biggl( E_{a_t} \biggl[ Q^2(s_t,a_t)\nabla_{\mu} \log \pi(a_t) \nabla_{\mu} \log \pi(a_t)^T \biggr]\\
& \quad - g^\mu(s_t) (g^\mu(s_t))^T \biggr).
\end{align*}
The sample-based estimate is the follow, where $s'\sim p(s_t)$ and $a'\sim \pi(a_t)$,
\begin{align*}
\hat{\Sigma}_{a,t}^{\mu,Q} &=  Q^2(s',a')\nabla_{\mu} \log \pi(a') \nabla_{\mu} \log \pi(a')^T \\
&\quad - g^\mu(s') (g^\mu(s'))^T.
\end{align*}
The sample-based estimate with $A(s,a)$-based estimator $\Sigma_{a,t}^{\mu,Q}$, i.e. $\phi(s)=V(s)$, is the same, except $Q(s,a)$ above is replaced by $A(s,a)$. In addition, an unbiased estimate for $ \Sigma_{a,t}^{\mu,Q} - \Sigma_{a,t}^{\mu,A}$ can be derived,
\begin{align*}
 \widehat{\Sigma_{a,t}^{\mu,Q} - \Sigma_{a,t}^{\mu,A}} &= \left(Q(s',a')^2-A(s',a')^2\right)\\
 &\quad \cdot \nabla_{\mu} \log \pi(a') \nabla_{\mu} \log \pi(a')^T.
\end{align*}
Importantly, $ \Sigma_{a,t}^{\mu,Q} - \Sigma_{a,t}^{\mu,A}$ is the variance reduction from using $V(s)$ as the state-dependent baseline, and $\Sigma_{a,t}^{\mu,A}$ is the extra variance reduction from using $\phi(s,a)=Q(s,a)$, the optimal variance reducing state-action-dependent baseline.

Lastly, analytic computation for the $\Sigma_\tau$ term is also tractable, since $\tau|s,a$ is a Gaussian, and the integrand is quadratic. However, computing this full Gaussian is computationally expensive, so instead we use the following form, which is still sufficiently low variance,
\begin{align*}
\Sigma_{\tau,t}^\mu &= E_{s_t} E_{a_t} \biggl( \nabla_{\mu} \log \pi(a_t) \nabla_{\mu} \log \pi(a_t)^T\\
&\quad \text{Var}_{\tau_{t+1}|s_t,a_t} \hat{Q}(s_t,a_t,\tau_{t+1})   \biggr)\\
&= E_{s_t} E_{a_t} \biggl( \nabla_{\mu} \log \pi(a_t)\nabla_{\mu} \log \pi(a_t)^T\\
& \quad \cdot  E_{\tau_{t+1}|s_t,a_t} \left(\hat{Q}(s_t,a_t,\tau_{t+1})^2-Q(s_t,a_t)^2\right)   \biggr).
\end{align*}
whose sample estimate is given with $s'\sim p(s_t)$, $a'\sim \pi(a_t)$ and $\tau'\sim \tau_{t+1}|s',a'$ as,
\begin{align*}
\hat{\Sigma}_{\tau,t}^\mu &= \nabla_{\mu} \log \pi(a_t)\nabla_{\mu} \log \pi(a_t)^T \\
&\quad \cdot \left(\hat{Q}(s_t,a_t,\tau_{t+1})^2-Q(s_t,a_t)^2\right).
\end{align*}

\subsection{LQG Experiments}
For experimental evaluation of variances in LQG, we used a simple 2D point mass system, whose dynamics and cost matrices are expressed as,
\begin{align*}
&A =
\begin{bmatrix}
   1 & 0 & dt & 0  \\
    0 & 1 & 0 & dt  \\
    0 & 0 & 1 & 0  \\
    0 & 0 & 0 & 1 
\end{bmatrix}, B = 
\begin{bmatrix}
  0 & 0   \\
    0 & 0  \\
   dt/m & 0   \\
    0 & dt/m   \\
\end{bmatrix} \\
& Q = \begin{bmatrix}
   q & 0 & 0 & 0  \\
    0 & q & 0 & 0  \\
    0 & 0 & q & 0  \\
    0 & 0 & 0 & q 
\end{bmatrix}, 
R = \begin{bmatrix}
   r & 0 & 0 & 0  \\
    0 & r & 0 & 0  \\
    0 & 0 & r & 0  \\
    0 & 0 & 0 & r 
\end{bmatrix},
\end{align*}
where we set $dt=0.05$, $m=1.0$, $q=1.0$, $r=0.01$. We set the initial state mean as $\mu^S_0 = [3.0, 4.0, 0.5, -0.5]$ and set the state variances as $\Sigma_0^S = \Sigma_t^{S|SA} = 0.0001I$. 

To initialize the time-varying open-loop policy, we draw the means from $\mu^A_t \sim \mathcal{N}(0, 0.3I)$ and set the variances as $\Sigma_t^A = 0.001I$. We choose time horizon $T=100$ for our experiments. To evaluate all variance terms, we use sample size $N=20000$ where Monte Carlo estimates are used.

\begin{figure*}[h]
  \centering
  %\begin{subfigure}
  %\begin{subfigure}[b]{0.99\textwidth}
  \includegraphics[width=0.97\textwidth]{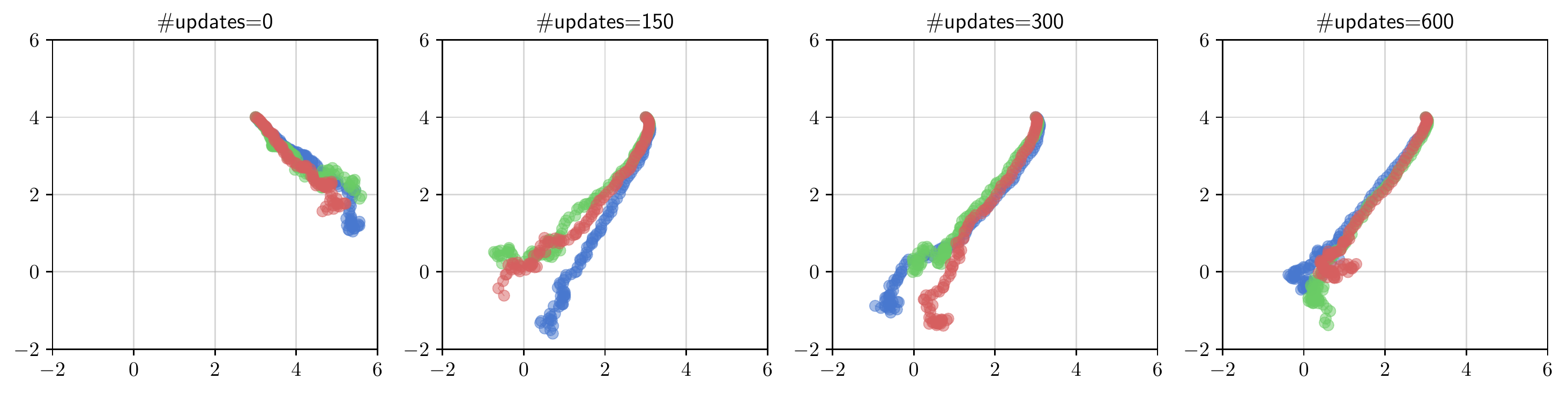}
  \includegraphics[width=0.99\textwidth]{out_variances_4row}
  %\end{subfigure}
  \caption{Evaluating the variance terms (Eq.~\ref{eq:var}) of the policy gradient estimator on a 2D point mass task (an LQG system) with finite horizon $T=100$. The figure includes Figure~\ref{fig:lqg_var} in addition to subplots visualizing on-policy trajectories at different stages of learning. The cost matrices encourage the point mass to reach and stay at the (0, 0) coordinate. Note that the task is challenging because the policy variances are not optimized and the policy uses open-loop controls. We include animated GIF visualizations of the variance terms and policy as learning progresses in the Supplementary Materials.} 
  \label{fig:lqg_var2}
\end{figure*}

We optimize the policy parameters to show how the variance decomposition changes as the policy is trained. We focus on analyzing the mean variances, so we only optimize $\mu^A_t$. Analytic gradients $g^\mu_t$ are used for gradient descent with learning rate $0.001$ and momentum $0.1$. The cost improves monotonically. Figure~\ref{fig:lqg_var2} complements Figure~\ref{fig:lqg_var} with additional visualizations on the policy at different stages of learning.

\section{Estimating variance terms}
\label{sec:est_var}
To empirically evaluate the variance terms, we use an unbiased single sample estimator for each of the three variance terms, which can be repeated to drive the measurement variance to an acceptable level.

First, we draw a batch of data according to $\pi$. Then, we select a random state $s$ from the batch and draw $a \sim \pi$. 

To estimate the first term, we draw $\tau, \tau' \sim \tau | s, a$ and the estimator is
\[ \left(\hat{A}(s, a, \tau)^2 - \hat{A}(s, a, \tau)\hat{A}(s, a, \tau')\right)\nabla \log \pi(a | s)^2. \]

We estimate the second term in two cases: $\phi(s, a) = 0$ and $\phi(s, a) = \hat{A}(s) := E_{a, \tau|s}\left[ \hat{A}(s, a, \tau) \right]$\footnote{When $\hat{A}(s, a, \tau) = \sum_t \gamma^tr_t$, then $\hat{A}(s) = V^\pi(s)$ a standard choice of a state-dependent control variate.}. When $\phi(s, a) = 0$, we draw $a'' \sim \pi$ and $\tau'' \sim \tau | s, a''$ and the estimator is
\begin{align*}
	\hat{A}(s, a, \tau)&\hat{A}(s, a, \tau')\nabla \log \pi(a | s)^2 \\
    &- \hat{A}(s, a, \tau)\hat{A}(s, a'', \tau'')\nabla \log \pi(a | s)\nabla \log \pi(a'' | s).
\end{align*}
When $\phi(s, a) = \hat{A}(s)$, we draw $a_1, a_2 \sim \pi$, $\tau_1 \sim \tau | s, a_1$, and $\tau_2 \sim \tau | s, a_2$ and the estimator is
\begin{align*}
	\biggr(&\hat{A}(s, a, \tau) - \hat{A}(s, a_1, \tau_1)\biggr)\left(\hat{A}(s, a, \tau') - \hat{A}(s, a_2, \tau_2)\right) \\
    &\times \nabla \log \pi(a | s)^2 \\
    &- \left(\hat{A}(s, a, \tau) - \hat{A}(s, a_1, \tau_1)\right) \\
    &\times \left(\hat{A}(s, a'', \tau'') - \hat{A}(s, a_2, \tau_2)\right) \\
    &\times \nabla \log \pi(a | s)\nabla \log \pi(a'' | s).
\end{align*}

Finally, the third term is
\begin{align*}
\Var_s &\left( E_{a|s} \left[ \hat{A}(s, a) \nabla \log \pi(a|s) \right] \right) \\
	&=  E_s \left[ E_{a|s} \left[ \hat{A}(s, a) \nabla \log \pi(a|s) \right]^2 \right] 
     - E_{s, a, \tau}[\hat{g}]^2 \\
    &\leq E_s \left[ E_{a|s} \left[ \hat{A}(s, a) \nabla \log \pi(a|s) \right]^2 \right]
\end{align*}
Computing an unbiased estimate of $E_{s, a, \tau}[\hat{g}]^2$ is straightforward, but it turns out to be small relative to all other terms, so we are satisfied with an upper bound. We can estimate the upper bound by
\[ \hat{A}(s, a, \tau)\hat{A}(s, a'', \tau'')\nabla \log \pi(a | s)\nabla \log \pi(a'' | s). \]

\section{Gradient Bias and Variance with IPG}
\label{sec:app_ipg}
Similarly to IPG~\citep{gu2017interpolated}, we consider a convex combination of estimators
\begin{align*}
	\hat{g}(s, a, \tau) =& \lambda \left( \hat{A}(s, a, \tau) - \phi(s, a) \right) \nabla \log \pi(a|s) \nonumber
    \\
    &+ \nabla \mathbb{E}_{a|s}\left[ \phi(s, a) \right],
\end{align*}
controlled by $\lambda \in [0, 1]$, where $\lambda = 1$ is unbiased and $\lambda = 0$ is low variance, but biased. Using the law of total variance, we can write the variance of the estimator as
\begin{align*}
    &\lambda^2 \mathbb{E}_s \left[ \Var_{a, \tau|s} \left( \left( \hat{A}(s, a, \tau) - \phi(s, a) \right) \nabla \log \pi(a|s) \right) \right] \\
    &+ \Var_s  \mathbb{E}_{a |s} \left[ \lambda \E_{\tau | s, a} \left[ \left( \hat{A}(s, a, \tau) \right] + (1 - \lambda)\phi(s, a) \right) \nabla \log \pi(a|s) \right].
\end{align*}
So when $\phi(s, a) \approx \E_{\tau | s, a} \left[ \hat{A}(s, a, \tau) \right]$, introducing $\lambda$ effectively reduces the variance by $\lambda^2.$ On the other hand, the bias introduced by $\lambda$ is
\[ (1 - \lambda) \E_{s, a} \left[ \left( \phi(s, a)  - \E_{\tau | s, a} \left[ \hat{A}(s, a, \tau) \right] \right) \nabla \log \pi(a|s) \right]. \]
Dynamically trading off between these quantities is a promising area of future research.

%\section{Horizon-aware Value Functions}
%\label{sec:app_horizon_aware}
%Suppose $\pi$ is stationary with respect to $p_0(s_0)$ (i.e., $s_t \sim p_0$), then the mean squared error of the value function is
%\begin{align*}
%    \E_{p_0(s)} &\left[ \left(\hat{V}(s) - \sum_t \gamma^t r_t \right)^2 \right] = \\
%    &\frac{1}{T+1} \sum_{l = 0}^T \E_{p_0(s)} \left[ \left(\hat{V}(s) - \sum_{t=0}^l \gamma^t r_t \right)^2 \right],
%\end{align*}
%due to the time limit. When the value function is unaware of time, this is minimized at
%\[ \hat{V}(s) = \frac{1}{T+1}\sum_{l = 0}^T \sum_{t = 0}^l \gamma^t r_t. \]

%\subsection{variance reduction with value function state baseline}
%Per state variance of the Monte Carlo policy gradient with no baselines.
%\begin{align*}
%\Sigma(s) =&~ {\color{red} E_{a} \Var_{\tau|a} \left( \hat{Q}(s, a, \tau) \nabla \log \pi(a|s) \right)} \\
%&+ {\color{blue}\Var_{a}  \biggr( Q(s, a) \nabla \log \pi(a|s) \biggr)}
%\end{align*}
%Per state variance of the state-action-dependent baseline.
%\begin{align*}
%\Sigma_{s,a}(s) =&~ {\color{red} E_{a} \Var_{\tau|a} \left( \hat{Q}(s, a, \tau) \nabla \log \pi(a|s) \right)} \\
%&+ \Var_{a}  \biggr( \left( Q(s, a) - \phi(s,a) \right) \nabla \log \pi(a|s) \biggr)
%\end{align*}
%Per state variance of the state-dependent baseline.
%\begin{align*}
%\Sigma_{s}(s) =&~ {\color{red} E_{a} \Var_{\tau|a} \left( \hat{Q}(s, a, \tau) \nabla \log \pi(a|s) \right)} \\
%&+ \Var_{a}  \biggr( \left( Q(s, a) - \psi(s) \right) \nabla \log \pi(a|s) \biggr) \\
%=& E_{a} \Var_{\tau|a} \left( \hat{Q}(s, a, \tau) \nabla \log \pi(a|s) \right) \\
%&+ \Var_{a}  \biggr( \left( Q(s, a) - \psi(s) \right) \nabla \log \pi(a|s) \biggr) 
%SG.02.01: I want to write down the variance reduction term when psi(s) = v(s). However, the expression is quite nasty. Let me know if there are better expressions.

%\end{align*}

\begin{figure*}[!ht]
  \centering
  \includegraphics[width=0.8\textwidth]{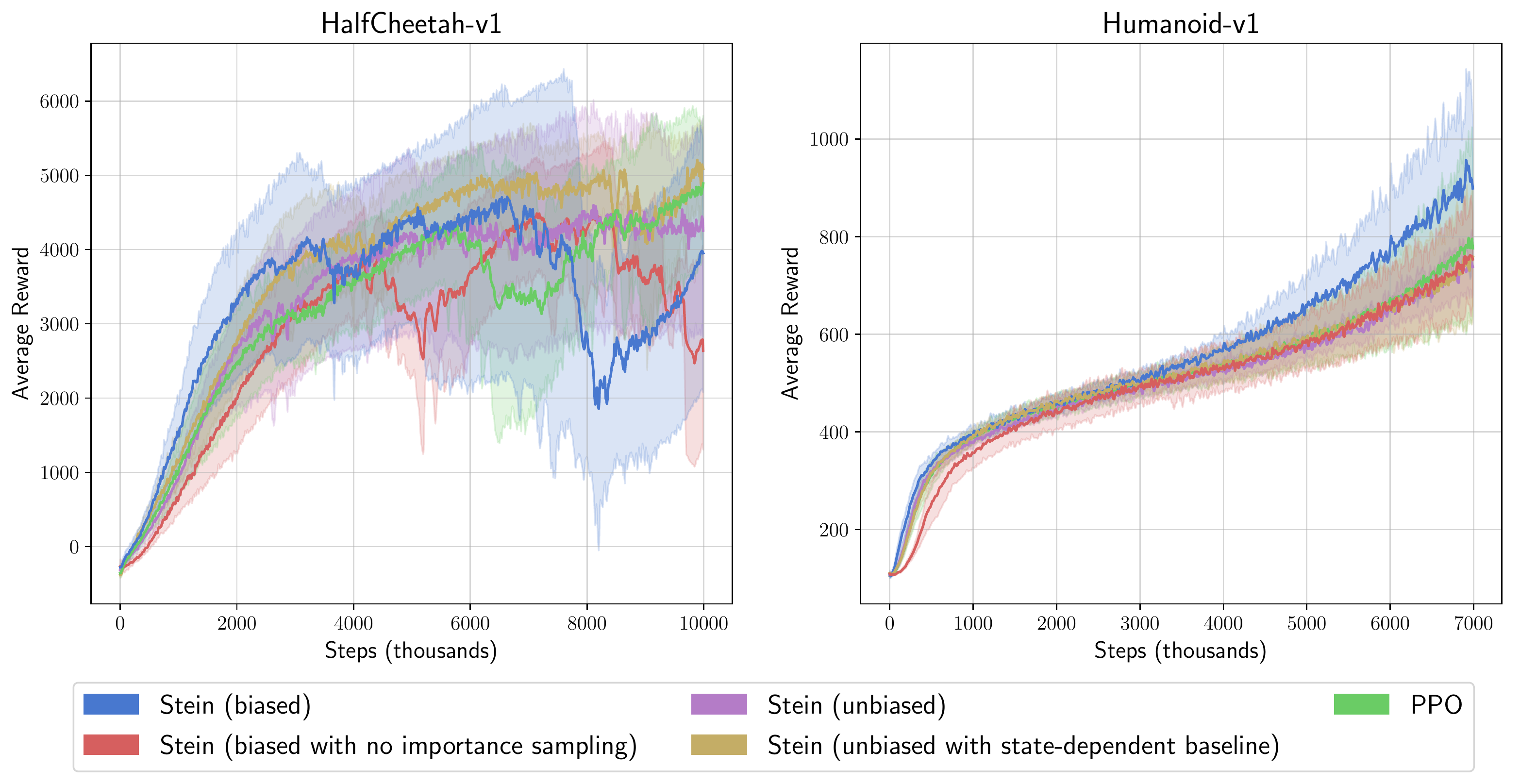}
  \caption{Evaluation of the (biased) Stein control variate state-action-dependent baseline, a biased variant without importance sampling for the bias correction term, an unbiased variant using a state-dependent baseline, an unbiased variant using only a state-dependent baseline, and PPO (implementations based on the code accompanying \citep{liu2018sample}). We plot mean episode reward and standard deviation intervals capped at the minimum and maximum across 5 random seeds. The batch size across all experiments was 10000.}
  \label{fig:stein}
  \vspace{-0.1in}
\end{figure*}

\begin{figure*}[!ht]
  \centering
  \includegraphics[width=0.9\textwidth]{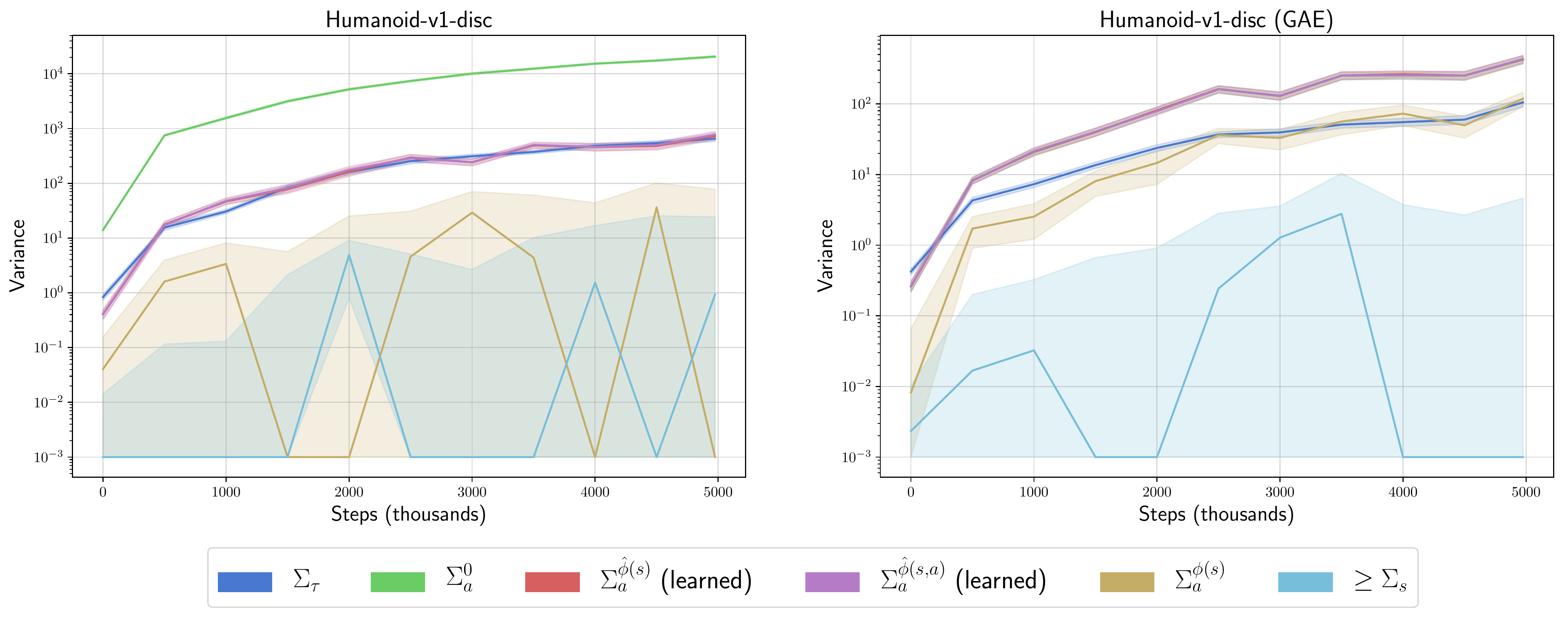}
  \caption{Evaluating the variance terms (Eq.~\ref{eq:var}) of the gradient estimator when $\hat{A}(s, a, \tau)$ is the discounted return (left) and GAE (right) with various baselines on Humanoid. The x-axis denotes the number of environment steps used for training. The policy is trained with TRPO and the horizon-aware value function. We set $\phi(s) = \mathbb{E}_{a}\left[\hat{A}(s, a)\right]$ and $\phi(s, a) = \hat{A}(s, a)$. The ``learned'' label in the legend indicates that a function approximator to $\phi$ was used instead of directly using $\phi$. Note that when using $\phi(s, a) = \hat{A}(s, a)$, $\Sigma_a^{\phi(s, a)}$ is $0$, so is not plotted. Since $\Sigma_s$ is small, we plot an upper bound on $\Sigma_s$. The upper and lower bands indicate two standard errors of the mean. In the left plot, lines for $\Sigma_a^{\hat{\phi}(s)}$ and $\Sigma_a^{\hat{\phi}(s, a)}$ overlap and in the right plot, lines for $\Sigma_a^0, \Sigma_a^{\hat{\phi}(s)}$, and $\Sigma_a^{\hat{\phi}(s, a)}$ overlap.}
  \label{fig:humanoid_disc_var}
\end{figure*}

\begin{figure*}[!ht]
  \centering
  \includegraphics[width=0.9\textwidth]{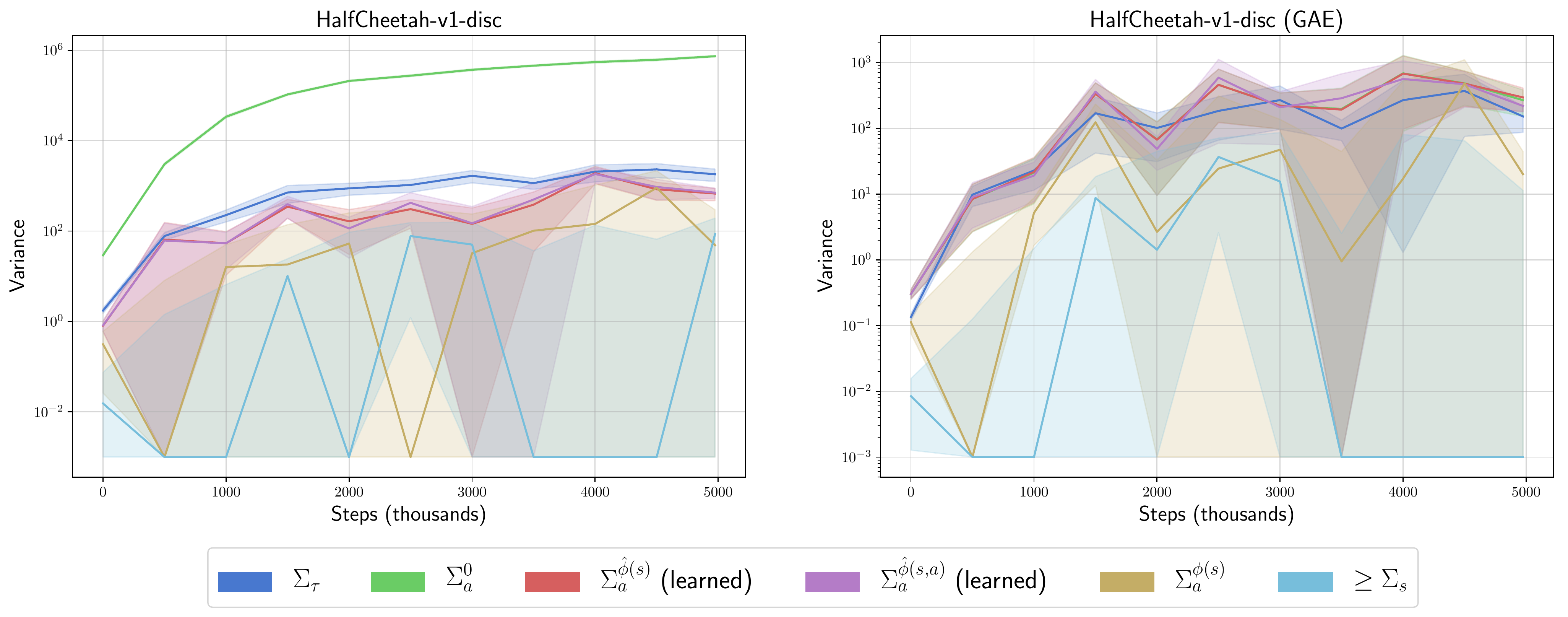}
  \caption{Evaluating the variance terms (Eq.~\ref{eq:var}) of the gradient estimator when $\hat{A}(s, a, \tau)$ is the discounted return (left) and GAE (right) with various baselines on HalfCheetah. The x-axis denotes the number of environment steps used for training. The policy is trained with TRPO and the horizon-aware value function. We set $\phi(s) = \mathbb{E}_{a}\left[\hat{A}(s, a)\right]$ and $\phi(s, a) = \hat{A}(s, a)$. The ``learned'' label in the legend indicates that a function approximator to $\phi$ was used instead of directly using $\phi$. Note that when using $\phi(s, a) = \hat{A}(s, a)$, $\Sigma_a^{\phi(s, a)}$ is $0$, so is not plotted. Since $\Sigma_s$ is small, we plot an upper bound on $\Sigma_s$. The upper and lower bands indicate two standard errors of the mean. In the left plot, lines for $\Sigma_a^{\hat{\phi}(s)}$ and $\Sigma_a^{\hat{\phi}(s, a)}$ overlap and in the right plot, lines for $\Sigma_a^0, \Sigma_a^{\hat{\phi}(s)}$, and $\Sigma_a^{\hat{\phi}(s, a)}$ overlap.}
  \label{fig:halfcheetah_disc_var}
\end{figure*}

\begin{figure*}[!ht]
  \centering
  \includegraphics[width=\textwidth]{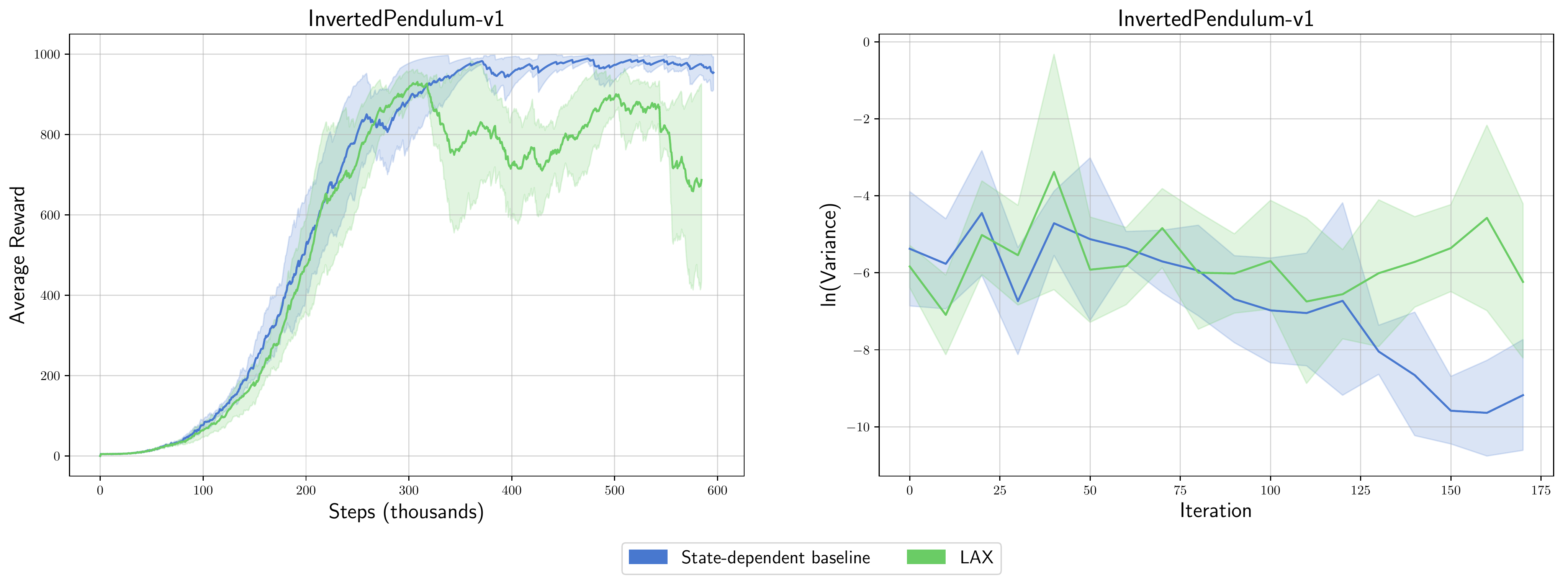}
  \caption{Evaluation of the state-action-dependent baseline (LAX) compared to a state-dependent baseline (Baseline) with the implementation from~\citet{grathwohl2018backpropagation} on the InvertedPendulum-v1 task. We plot episode reward (left) and log variance of the policy gradient estimator (right) averaged across 5 randomly seeded training runs. The error intervals are a single standard deviation capped at the min and max across the 5 runs. We used a batch size of 2500 steps.}
  \label{fig:bttv}
\end{figure*}

\end{document}